\documentclass[hidelinks,onefignum,onetabnum]{siamart250211}

\usepackage{lipsum}
\usepackage{amsfonts}
\usepackage{graphicx}
\usepackage{epstopdf}
\usepackage{algorithmic}
\ifpdf
  \DeclareGraphicsExtensions{.eps,.pdf,.png,.jpg}
\else
  \DeclareGraphicsExtensions{.eps}
\fi

\newsiamremark{remark}{Remark}
\newsiamremark{hypothesis}{Hypothesis}
\crefname{hypothesis}{Hypothesis}{Hypotheses}
\newsiamthm{claim}{Claim}
\newsiamremark{fact}{Fact}
\crefname{fact}{Fact}{Facts}

\headers{Unlearning Noise in PINNs}{Y. Chen, Y. Chen, W. Guo, and X. Zhong}

\title{Unlearning Noise in PINNs: A Selective Pruning Framework for PDE Inverse Problems
 \thanks{Submitted to the editors March 2026.}
\funding{The first and fourth authors were partially supported by National Key R\&D Program of China 2024YFA1012302 and the NSFC Grant 12272347.}}
\author{
Yongsheng Chen\thanks{School of Mathematical Sciences, Zhejiang University, Hangzhou 310027, China. \tt{22035024@zju.edu.cn}}
\and Yong Chen\thanks{Department of Biostatistics, Epidemiology and Informatics, University of Pennsylvania, Philadelphia, PA 19104, USA. \tt{ychen123@pennmedicine.upenn.edu}}
\and Wei Guo\thanks{Department of Mathematics and Statistics, Texas Tech University, Lubbock, TX 70409, USA.
  \tt{weimath.guo@ttu.edu}}
\and Xinghui Zhong\thanks{School of Mathematical Sciences, Zhejiang University, Hangzhou 310027, China. \tt{zhongxh@zju.edu.cn}}
}

\usepackage{amsopn}
\usepackage{multirow}
\usepackage{subfigure}
\usepackage{booktabs,tabularx} %

\newcommand{\bxi}{\boldsymbol{\xi}}

\newcommand{\calD}{\mathcal{D}}
\newcommand{\calR}{\mathcal{R}}

\ifpdf
\hypersetup{
  pdftitle={Unlearning Noise in PINNs},
  pdfauthor={Y. Chen, et al.}
}
\fi

\begin{document}

\maketitle

\begin{abstract}
Physics-informed neural networks (PINNs) provide a promising framework for solving inverse problems governed by partial differential equations (PDEs) by integrating observational data and physical constraints in a unified optimization objective. However, the ill-posed nature of PDE inverse problems makes them highly sensitive to noise. Even a small fraction of corrupted observations can distort internal neural representations, severely impairing accuracy and destabilizing training.
Motivated by recent advances in machine unlearning and structured network pruning, we propose P-PINN, a selective pruning framework designed to unlearn the influence of corrupted data in a pretrained PINN. Specifically, starting from a PINN trained on the full dataset, P-PINN evaluates a joint residual--data fidelity indicator,  a weighted combination of data misfit and PDE residuals, to partition the training set into reliable and corrupted subsets. Next, we introduce a bias-based neuron importance measure that quantifies directional activation discrepancies between the two subsets, identifying neurons whose representations are predominantly driven by corrupted samples. Building on this, an iterative pruning strategy then removes noise-sensitive neurons layer by layer. The resulting pruned network is fine-tuned on the reliable data subject to the original PDE constraints, acting as a lightweight post-processing stage rather than a complete retraining. Numerical experiments on extensive PDE inverse-problem benchmarks demonstrate that  P-PINN substantially improves robustness, accuracy, and training stability under noisy conditions, achieving up to a 96.6\% reduction in relative error compared with baseline PINNs. These results indicate that activation-level post hoc pruning is a promising mechanism for enhancing the reliability of physics-informed learning in noise-contaminated settings. An implementation is available at \url{https://github.com/chenyongssss/P_PINN}.
\end{abstract}

\begin{keywords}
PINN; Inverse PDE problems; Selective pruning; Noisy data; Machine unlearning;
\end{keywords}

\begin{MSCcodes}
68T07; 65N21; 65M32
\end{MSCcodes}

\section{Introduction}
Inverse problems constrained by partial differential equations (PDEs) play a central role in a broad range of scientific and engineering disciplines, including subsurface flow, thermal transport, structural mechanics, and fluid dynamics. The objective is to infer unknown coefficients, source terms, or boundary data from indirect and noisy observations of the system state. However, such problems are typically ill-posed: the parameter-to-observable map may be non-invertible or severely ill-conditioned, and the available data are often sparse, irregular, and corrupted by noise. Robust numerical methods must stabilize the inversion by incorporating prior information, regularity assumptions, and the governing PDE. Traditional approaches to PDE inverse problems, including variational optimization \cite{tarantola2005inverse,hansen1998rank}, Bayesian inference \cite{stuart2010inverse,kaipio2006statistical}, adjoint-based methods \cite{gunzburger2002perspectives,giles2000introduction}, and regularization techniques \cite{willoughby1979solutions,rudin1992nonlinear,daubechies2004iterative}, provide theoretically sound frameworks but struggle with high-dimensional parameter spaces, complex geometries, scarce and noisy observations, and intensive computational demands \cite{benning2018modern,arridge2019solving,isakov2017inverse}.

Recent machine-learning-based surrogates have been proposed to alleviate these challenges by approximating forward and inverse maps in a data-driven manner while incorporating physical structure. Deep operator networks such as DeepONets, Fourier neural operators, and graph-based models learn mappings between function spaces, for example from coefficients to solutions \cite{lu2021learning,li2021fourier,anandkumar2020neural}. Physically structured generative models and diffusion-based frameworks have been developed for probabilistic inversion, enabling approximate posterior sampling in high-dimensional settings \cite{yang2020physics,zhu2018bayesian,bar2019unsupervised,huang2024diffusionpde}. Among these developments, physics-informed neural networks (PINNs) have emerged as a particularly versatile tool for PDE-constrained learning and inverse problems \cite{raissi2019physics,mao2020physics,karniadakis2021physics}. A PINN approximates the solution field using a neural network surrogate, trained by minimizing a composite loss function involving  data misfit and PDE residual terms \cite{lu2021deepxde,jagtap2020conservative,wang2021understanding}. By naturally enforcing physical constraints, PINNs allow for the simultaneous recovery of unknown parameters and state fields. The framework has been further extended to incorporate adaptive weighting, domain decomposition, and uncertainty quantification \cite{wang2022and,xiang2022self,yang2021b,zhu2019physics}.

Despite these advances, baseline PINNs remain highly vulnerable to observational noise in practice. Measurement errors, stemming from sensor limitations or environmental fluctuations, inevitably contaminate training data. When such noise is significant, the data misfit terms in the PINN loss can drive internal activations to fit noise rather than the underlying physical signal. This effect is particularly detrimental in inverse problems, where reliable data are crucial for constraining ill-posed solution manifolds and reducing uncertainty \cite{ji2021stiff,lu2023remaining,linka2022bayesian}. Noise-induced distortions in the learned representation may lead to biased parameter estimates, degraded predictive accuracy, and unstable training dynamics, severely limiting the applicability of PINNs in realistic settings with imperfect measurements.

Several strategies have been proposed to improve noise robustness in data-driven PDE frameworks. Bayesian PINNs (B-PINNs) replace deterministic surrogates with Bayesian neural networks, enabling explicit modeling of both aleatoric and epistemic uncertainty and providing enhanced stability in high-noise regimes \cite{yang2021b,stock2023bayesian}. Dropout-based PINNs provide approximate uncertainty estimates via Monte Carlo dropout at inference time \cite{gal2016dropout,zhang2019quantifying}. Robust losses and regularization schemes modify the training objective or constrain parameter norms to reduce sensitivity to outliers and corrupted samples \cite{hua2023physics,karniadakis2021physics,wu2023comprehensive,xiang2022self,rojas2024robust}. While effective, these approaches primarily act at the level of optimization, loss weighting, or posterior modeling. They do not directly address the fundamental challenge: \emph{noise-corrupted observations can imprint biased internal representations within the network architecture itself}.

Concurrently, a growing line of research on machine unlearning has emerged, motivated by the need to eliminate the undesired influence of specific data points from trained models \cite{cao2015towards,bourtoule2021machine,graves2021amnesiac}. Early work relied on influence functions to approximate the impact of individual samples on model parameters \cite{koh2017understanding}, while subsequent methods such as DeltaGrad and SISA improve scalability by storing gradients or partitioning the dataset into shards to enable efficient updates after data removal \cite{wu2020deltagrad,bourtoule2021machine}. More recently, pruning-based unlearning methodologies are developed to identify and remove neurons or channels whose activations are strongly associated with the data intended for removal, thereby excising data-specific information through structured sparsification rather than full retraining \cite{ma2022learn,golatkar2020eternal,bragagnolo2021role}. Further, interpretability techniques based on directional activation discrepancies and bias scores provide effective tools to detect neurons that respond disproportionately to particular subsets of inputs \cite{kim2018interpretability,turner2023activation}. These developments suggest a promising path toward robust, data-aware modification of neural network architectures, yet their integration into PINNs and, more broadly, into scientific machine learning remains largely unexplored.

This work proposes a physics-guided selective pruning framework for PINNs, termed P-PINN, that reinterprets noise mitigation as a structured unlearning task. Starting from a standard PINN trained on noisy observations, we construct a data-driven partition of the training set into \emph{retained} (reliable) and \emph{forget} (suspect) subsets using a composite indicator that combines data misfit with PDE residual information, thereby identifying observations that are inconsistent with both the learned surrogate and the governing equations. We then introduce a bias-based neuron importance score that quantifies directional discrepancies in neuron activations between the two subsets. Using this metric, neurons whose activations are predominantly influenced by the \emph{forget} dataset are iteratively pruned, with importance scores recomputed after each pruning step to mitigate compensatory effects. The resulting pruned network is fine-tuned exclusively on the \emph{retained} dataset, yielding a network that preserves information from trustworthy measurements while effectively ``unlearning'' noise-driven artifacts.


Our main contributions are summarized as follows:
\begin{itemize}
  \item We define a physics-guided residual--data-fidelity indicator that combines data misfit and PDE residuals to separate reliable observations from corrupted samples in a fully data-driven manner.
  \item We introduce a bias-based neuron importance score that leverages directional activation discrepancies between retained and forget subsets to identify neurons whose activations are predominantly driven by corrupted data.
  \item We integrate these components into P-PINN, an activation-level unlearning framework comprising partitioning, iterative pruning with score recomputation, and selective fine-tuning on the retained dataset, which yields substantial robustness and accuracy gains (up to a $96.6\%$ reduction in relative error) compared to baseline models across diverse PDE benchmarks.
\end{itemize}

The remainder of the paper is organized as follows. In Section~\ref{sec:inverse-pinn} we formalize the class of PDE-constrained inverse and data assimilation problems considered in this work and summarize the standard PINN formulation. Section~\ref{sec:pruning} presents the proposed P-PINN framework, including the residual--data-fidelity metric for data partitioning, the bias-based neuron importance measure, and the iterative pruning and fine-tuning procedure. Section~\ref{sec:numerics-setup} describes the benchmark problems, numerical setups, and implementation details. Section~\ref{sec:numerical-results} reports numerical results and ablation studies that quantify the impact of the proposed components on accuracy, robustness, and convergence. Section~\ref{sec:conclusion} concludes with a discussion of limitations and directions for future research.

\vspace*{-0.25\baselineskip}
\section{Inverse PDE Problems and PINNs}
\label{sec:inverse-pinn}

In this section, we formalize the class of deterministic PDE inverse problems considered in this work, including parameter inversion and data assimilation, and summarize the baseline PINN formulation within a unified framework. This establishes the mathematical foundation for the selective pruning methodology developed in Section~\ref{sec:pruning}. Throughout, we assume that the observational data are contaminated by additive noise.
\subsection{PDE-constrained inverse problems}
\label{subsec:inverse-pde}

Let $\Omega \subset \mathbb{R}^{d_x}$ denote a bounded spatial domain and $I = [0,T] \subset \mathbb{R}$ a time interval. We denote a generic space--time location by
\[
\boldsymbol{\xi} = (\mathbf{x},t) \in \Omega \times I.
\]
For stationary problems, we define $I = \{0\}$. We consider an unknown state field
$u^\star : \Omega \times I \to \mathbb{R}^{d_u}$
and an unknown parameter (or coefficient) field
$\gamma^\star : \Omega \times I \to \mathbb{R}^{d_\gamma}$
which together satisfy a PDE given by
\begin{equation}
\mathcal{L}_{\gamma^\star}\big(u^\star\big)(\boldsymbol{\xi}) = f(\boldsymbol{\xi}), 
\quad \boldsymbol{\xi} \in \Omega \times I,
\label{eq:forward-pde}
\end{equation}
subject to boundary and initial conditions
\begin{equation}
\mathcal{B}_{\gamma^\star}\big(u^\star\big)(\boldsymbol{\xi}) = g(\boldsymbol{\xi}),
\quad \boldsymbol{\xi} \in \partial \Omega \times I,
\qquad
\mathcal{I}_{\gamma^\star}\big(u^\star\big)(\mathbf{x}) = h(\mathbf{x}),
\quad \mathbf{x} \in \Omega,
\label{eq:boundary-initial}
\end{equation}
where $\mathcal{L}_{\gamma^\star}$ denotes the differential operator of the PDE, $\mathcal{B}_{\gamma^\star}$ encodes boundary conditions, and $\mathcal{I}_{\gamma^\star}$ encodes initial conditions for time-dependent problems.

We assume access to $N$ noisy observations of the state,
\begin{equation}
\mathcal{D}
:=
\{(\boldsymbol{\xi}_i,\bar{u}_i)\}_{i=1}^N,\quad \bar{u}_i = u^\star (\boldsymbol{\xi}_i) + \eta_i,
\label{eq:observations}
\end{equation}
where $\xi_i=(x_i,t_i)$ are sampling locations and $\eta_i$ denotes measurement noise.
In our numerical studies, we model $\eta_i$ as independent Gaussian perturbations
\(
\eta_i\sim\mathcal{N}(0,\sigma_i^2),
\)
allowing heterogeneous noise levels across observations. We emphasize, however, that the proposed framework treats $\{\eta_i\}$ as generic perturbations and does not rely on the Gaussian assumption.

The objective of the PDE-constrained inverse problem is to recover the unknown quantities $\gamma^\star$ and $u^\star$ from the noisy observations $\mathcal{D}$, subject to the physical constraints~\eqref{eq:forward-pde}--\eqref{eq:boundary-initial}. We focus on two classes of inverse problems:
\begin{itemize}
\item \emph{Parameter inversion:} the primary unknown is $\gamma^\star$ (e.g., spatially varying diffusivity, source terms, or material coefficients), with $u^\star$ being of secondary interest. The observations typically consist of sparse and noisy pointwise measurements of $u^\star$.
\item \emph{Data assimilation:} the parameter $\gamma^\star$ is known or fixed, while the goal is to reconstruct the state $u^\star$ from partial, noisy observations in space and time. Examples include reconstructing transient temperature fields, wave fields, or incompressible velocity fields.
\end{itemize}
Both problem classes are often severely ill-posed: the mapping from $(\gamma,u)$ to observations may be non-invertible, and small perturbations in data can induce large variations in the inferred quantities. Incorporating the PDE constraints into the inversion procedure is therefore essential to stabilize the problem and obtain physically meaningful reconstructions.

\subsection{PINNs for inverse problems with noisy data}
\label{subsec:pinn-formulation}

PINNs approximate the state $u^\star$ and, where applicable, the parameters $\gamma^\star$ using trainable neural networks while enforcing the governing PDE~\eqref{eq:forward-pde} and the associated boundary/initial conditions~\eqref{eq:boundary-initial} through the training objective. We briefly summarize the formulation below.

We approximate the state $u^\star$ by a neural network $u_\theta : \Omega \times I \to \mathbb{R}^{d_u}$,
parameterized by weights and biases $\theta \in \mathbb{R}^{d_\theta}$. For parameter inversion, we introduce a second trainable network (or trainable variables) $\gamma_\phi : \Omega \times I \to \mathbb{R}^{d_\gamma},$
with parameters $\phi \in \mathbb{R}^{d_\phi}$. Depending on the specific application, $\gamma_\phi$ may represent global scalars, a spatially varying field, or a spatiotemporal function.

The residual of the governing PDE at a point $\boldsymbol{\xi} \in \Omega \times I$ is defined as
\vspace*{-0.7\baselineskip}
\begin{equation}
\mathcal{R}_{\text{PDE}}(u_\theta,\gamma_\phi)(\boldsymbol{\xi})
=
\mathcal{L}_{\gamma_\phi}\big(u_\theta\big)(\boldsymbol{\xi}) - f(\boldsymbol{\xi}),
\label{eq:pde-residual}
\vspace*{-0.7\baselineskip}
\end{equation}
where the derivatives of $u_\theta$ with respect to $\mathbf{x}$ and $t$ are computed via automatic differentiation. Similarly, boundary and initial residuals are given by
\vspace*{-0.7\baselineskip}
\begin{align}
\mathcal{R}_{\text{BC}}(u_\theta,\gamma_\phi)(\boldsymbol{\xi})
&=
\mathcal{B}_{\gamma_\phi}\big(u_\theta\big)(\boldsymbol{\xi}) - g(\boldsymbol{\xi}),
\quad \boldsymbol{\xi} \in \partial \Omega \times I,
\label{eq:bc-residual}
\\
\mathcal{R}_{\text{IC}}(u_\theta,\gamma_\phi)(\mathbf{x})
&=
\mathcal{I}_{\gamma_\phi}\big(u_\theta\big)(\mathbf{x}) - h(\mathbf{x}),
\quad \mathbf{x} \in \Omega.
\label{eq:ic-residual}
\vspace*{-0.7\baselineskip}
\end{align}
In practice, these residuals are evaluated at sets of collocation points sampled from the space-time domain, boundary, and initial time slice. In particular, given the noisy observations $\mathcal{D}$, the standard PINN loss function is a weighted sum of the data misfit term and the PDE residuals:
\vspace*{-0.7\baselineskip}
\begin{equation}
\mathcal{J}(\theta,\phi)
=
\lambda_{\text{data}} \, \mathcal{J}_{\text{data}}(\theta)
+
\lambda_{\text{PDE}} \, \mathcal{J}_{\text{PDE}}(\theta,\phi)
+
\lambda_{\text{BC}} \, \mathcal{J}_{\text{BC}}(\theta,\phi)
+
\lambda_{\text{IC}} \, \mathcal{J}_{\text{IC}}(\theta,\phi),
\label{eq:full-loss}
\vspace*{-0.7\baselineskip}
\end{equation}
where the user-specified weights $\lambda_{\text{data}},\lambda_{\text{PDE}},\lambda_{\text{BC}},\lambda_{\text{IC}} > 0$ balance the competing objectives. The individual loss terms are given by
\vspace*{-0.7\baselineskip}
\begin{align}
\mathcal{J}_{\text{data}}(\theta)
&=
\frac{1}{N}
\sum_{i=1}^N
\big\|
u_\theta(\boldsymbol{\xi}_i) - \bar{u}_i
\big\|_2^2,
\label{eq:data-loss}
\\
\mathcal{J}_{\text{PDE}}(\theta,\phi)
&=
\frac{1}{N_{\text{PDE}}}
\sum_{j=1}^{N_{\text{PDE}}}
\big\|
\mathcal{R}_{\text{PDE}}(u_\theta,\gamma_\phi)(\boldsymbol{\xi}_j^{\text{PDE}})
\big\|_2^2,
\label{eq:pde-loss}
\\
\mathcal{J}_{\text{BC}}(\theta,\phi)
&=
\frac{1}{N_{\text{BC}}}
\sum_{k=1}^{N_{\text{BC}}}
\big\|
\mathcal{R}_{\text{BC}}(u_\theta,\gamma_\phi)(\boldsymbol{\xi}_k^{\text{BC}})
\big\|_2^2,
\label{eq:bc-loss}
\\
\mathcal{J}_{\text{IC}}(\theta,\phi)
&=
\frac{1}{N_{\text{IC}}}
\sum_{m=1}^{N_{\text{IC}}}
\big\|
\mathcal{R}_{\text{IC}}(u_\theta,\gamma_\phi)(\boldsymbol{\xi}_m^{\text{IC}})
\big\|_2^2.
\label{eq:ic-loss}
\end{align}
Here $\{\boldsymbol{\xi}_j^{\text{PDE}}\}_{j=1}^{N_{\text{PDE}}}$, $\{\boldsymbol{\xi}_k^{\text{BC}}\}_{k=1}^{N_{\text{BC}}}$, and $\{\boldsymbol{\xi}_m^{\text{IC}}\}_{m=1}^{N_{\text{IC}}}$ denote the collocation points for enforcing the PDE, boundary, and initial conditions, respectively. For stationary problems, the term $\mathcal{J}_{\text{IC}}$ is omitted.

The training of the PINN amounts to solving the optimization problem
\vspace*{-0.7\baselineskip}
\begin{equation}
(\theta^\star,\phi^\star)
=
\arg\min_{\theta,\phi} \, \mathcal{J}(\theta,\phi)
\label{eq:pinn-training}
\vspace*{-0.7\baselineskip}
\end{equation}
via stochastic gradient-based algorithms. Upon convergence, the reconstructed fields are given by $u_{\theta^\star} \approx u^\star$ and $\gamma_{\phi^\star} \approx \gamma^\star$.

\begin{figure}[t]
  \centering
  \includegraphics[width=\linewidth]{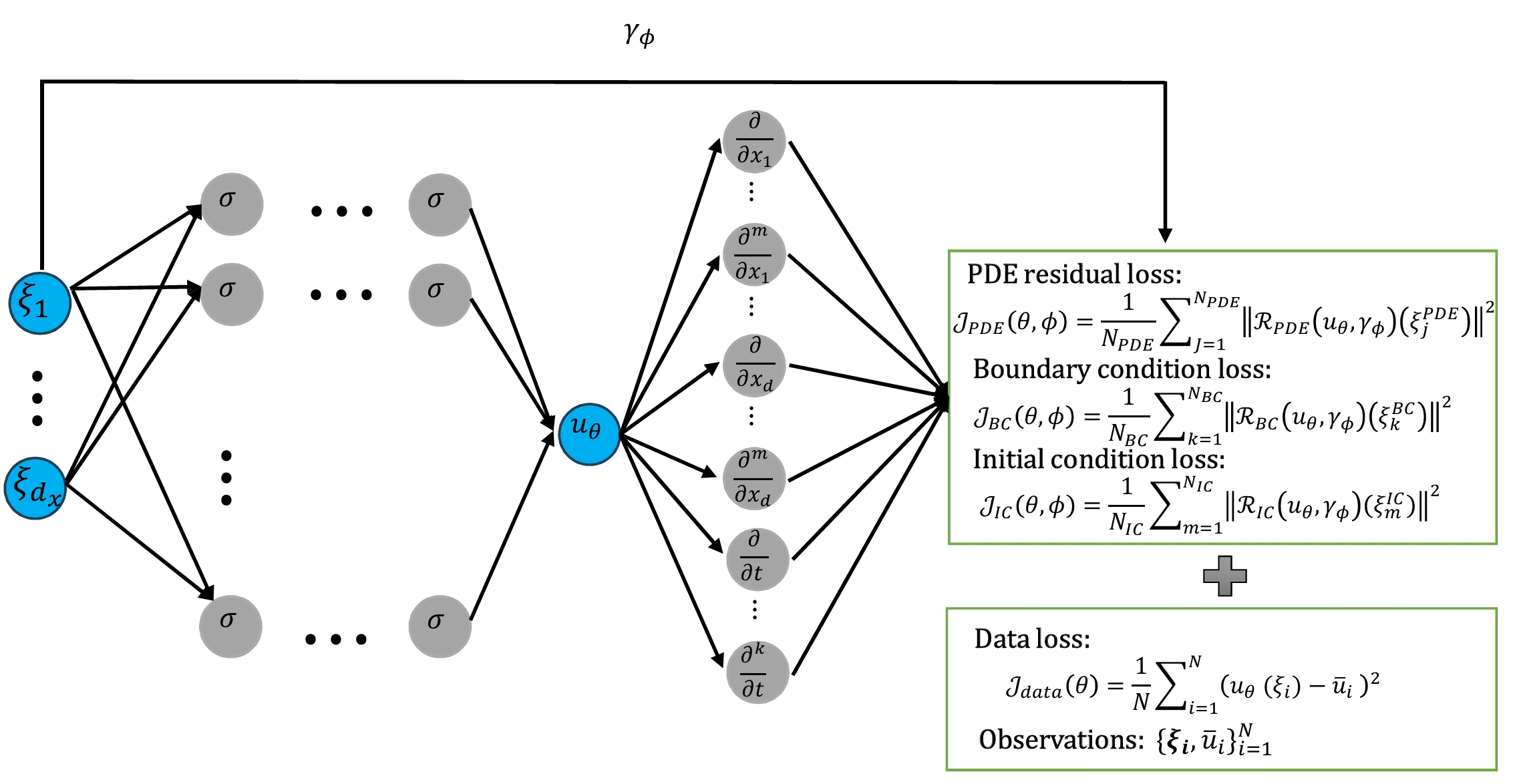}
  \caption{Schematic representation of the PINN formulation used in this work.}
  \label{fig:pinn-schematic}
\end{figure}

This formulation is sufficiently general to encompass both parameter inversion and data assimilation tasks. In parameter inversion, the primary objective is to identify the unknown coefficients $\gamma_\phi$; in data assimilation, $\gamma$ is typically fixed, and $\gamma_\phi$ is excluded from the optimization. A schematic illustration of this standard PINN formulation, highlighting the network architecture and the composite training objective, is shown in Figure~\ref{fig:pinn-schematic}. 

\subsection{Challenges under noisy observations and motivation}
\label{subsec:challenges-motivation}

In the low-noise regime, the composite PINN loss~\eqref{eq:full-loss} can effectively balance data fidelity with physical consistency, yielding accurate reconstructions. However, as the noise level increases or when the data quality is heterogeneous, the data misfit term~\eqref{eq:data-loss} may dominate the training dynamics for certain samples. Inverse problems are particularly sensitive to this effect: noise-corrupted measurements can drive the learned state $u_\theta$ and parameters $\gamma_\phi$ away from the physically consistent solution manifold, while the PDE residual terms may not be sufficient to counteract this bias without careful loss-weight tuning or sophisticated regularization strategies.

From a representation learning perspective, heavily corrupted observations 
tend to imprint noise-induced patterns on internal activations, especially in deeper 
layers \cite{kim2018interpretability,turner2023activation}. These distorted activations 
can propagate through training, resulting in biased reconstructions and poor 
generalization. Conventional remedies, such as robust loss functions 
\cite{hua2023physics,rojas2024robust}, adaptive weighting strategies 
\cite{xiang2022self}, or Bayesian formulations 
\cite{yang2021b,stock2023bayesian}, primarily operate at the level of the objective 
function or uncertainty modeling. They do not explicitly identify and remove the 
specific internal neurons that are strongly associated with noisy data. This 
limitation motivates the development of activation-level interventions. In 
Section~\ref{sec:pruning}, we introduce a selective pruning framework designed to 
detect and remove these noise-sensitive components, thereby enhancing the 
robustness and accuracy of the baseline PINN solver.

\section{A Selective Pruning Framework for Noisy PINNs}
\label{sec:pruning}

We propose P-PINN, a selective pruning framework that serves as a post hoc, activation-level unlearning mechanism for noisy PINNs. Starting from a baseline PINN $(u_{\theta^0},\gamma_{\phi^0})$ trained on the full noisy dataset by minimizing~\eqref{eq:full-loss}, P-PINN consists of: (i) a residual--data-fidelity partition of observations into retained and forget subsets, (ii) a bias-based neuron-importance score computed from activation discrepancies across the two subsets, and (iii) iterative pruning with score recomputation followed by selective fine-tuning on the retained subset. The resulting network suppresses the influence of severely corrupted observations while preserving information supported by reliable data.


\subsection{Data-driven partitioning via residual--fidelity metrics}
\label{subsec:partition}

Let $(u_{\theta^0},\gamma_{\phi^0})$ be the baseline PINN on noisy data. We assign each observation $(\bxi_i, \bar{u}_i) \in \calD$ a scalar score quantifying its compatibility with both the current model prediction and the governing PDE. We define the pointwise data misfit $r^{\text{data}}_i$ and PDE residual $r^{\text{PDE}}_i$ as 
\vspace*{-0.2\baselineskip}
\begin{equation}\label{eq:point}
    r^{\text{data}}_i = \big\| u_{\theta^0}(\bxi_i) - \bar{u}_i \big\|_2, \qquad
    r^{\text{PDE}}_i = \big\| \calR_{\text{PDE}}(u_{\theta^0},\gamma_{\phi^0})(\bxi_i) \big\|_2.
\vspace*{-0.2\baselineskip}
\end{equation}
The composite fidelity--residual score $M_i$ is given by
\vspace*{-0.2\baselineskip}
\begin{equation}
    M_i = \alpha_{\text{data}} \, r^{\text{data}}_i + \alpha_{\text{PDE}} \, r^{\text{PDE}}_i, \quad i = 1,\dots,N,
    \label{eq:composite-metric}
\vspace*{-0.2\baselineskip}
\end{equation}
where $\alpha_{\text{data}}, \alpha_{\text{PDE}} > 0$ are user-specified weights. The first term in~\eqref{eq:composite-metric} penalizes large discrepancies between the model predictions and the noisy data, while the second term penalizes local violations of the PDE. Observations that are both hard to fit and inconsistent with the PDE are thus assigned large scores.

Once scores $\{M_i\}_{i=1}^N$ are computed, we choose a threshold $\tau > 0$ and define the retained and forget index sets by
\begin{equation}
\mathcal{I}_{\text{retain}}
=
\{ i \in \{1,\dots,N\} : M_i < \tau \},
\qquad
\mathcal{I}_{\text{forget}}
=
\{ i \in \{1,\dots,N\} : M_i \ge \tau \}.
\label{eq:retain-forget-sets}
\end{equation}
The corresponding retained and forget datasets are defined as 
\[
\mathcal{D}_{\text{retain}}
=
\{(\boldsymbol{\xi}_i,\bar{u}_i) : i \in \mathcal{I}_{\text{retain}}\},
\qquad
\mathcal{D}_{\text{forget}}
=
\{(\boldsymbol{\xi}_i,\bar{u}_i) : i \in \mathcal{I}_{\text{forget}}\}.
\]

In our numerical experiments, we select $\tau$ via a prescribed retention ratio. Specifically, for a given $\rho \in (0,1)$, we set $\tau$ to the empirical $\rho$-quantile of $\{M_i\}_{i=1}^N$ so that approximately $\rho N$ observations are retained. We find that the P-PINN framework is  robust with respect to this choice: even when only a portion of the highly corrupted observations is assigned to $\mathcal{D}_{\text{forget}}$, or when a relatively aggressive fraction of data is removed, the resulting P-PINN consistently outperforms the baseline PINN trained on the full, unpartitioned dataset. In the simulations, we employ moderate retention ratios (e.g., $40\%$--$80\%$), which strike a balance between information preservation and noise removal and already yield reliable improvements across all benchmark problems.

The partition~\eqref{eq:retain-forget-sets} serves two roles. First, it provides a physics-informed mechanism to identify observations that are dominated by noise or model mismatch. Second, it induces two data distributions that will be used to measure the bias of neuron activations in the network, thereby guiding the subsequent pruning stage.

\subsection{Bias-based neuron importance and pruning criteria}
\label{subsec:bias-score}

We construct an importance metric that identifies neurons whose activations are disproportionately influenced by  $\mathcal{D}_{\text{forget}}$. The central idea is to quantify, for each neuron, the directional difference between its mean activations on the forget and retained subsets. We designate neurons with large differences as noise-sensitive.

We consider a standard fully connected multilayer-perceptron (MLP) PINN architecture with $L$ hidden layers. For $\ell = 1,\dots,L$, let $N_\ell$ denote the number of neurons in layer $\ell$. We denote by
$
a^{(\ell)}(\boldsymbol{\xi}) \in \mathbb{R}^{N_\ell}
$
the post-activation vector (i.e., after applying the nonlinear activation function) at layer $\ell$ for input $\boldsymbol{\xi}$, and let $a^{(\ell)}_n(\boldsymbol{\xi})$ denote its $n$th component, $n=1,\dots,N_\ell$.

For each layer $\ell$ and neuron index $n$ (neuron $(\ell,n)$), we define the average activation over the retained and forget subsets, respectively:
\vspace*{-0.4\baselineskip}
\begin{equation}
\mu^{(\ell)}_{\text{retain},n}
=
\frac{1}{|\mathcal{I}_{\text{retain}}|}
\sum_{i \in \mathcal{I}_{\text{retain}}}
a^{(\ell)}_n(\boldsymbol{\xi}_i),
\qquad
\mu^{(\ell)}_{\text{forget},n}
=
\frac{1}{|\mathcal{I}_{\text{forget}}|}
\sum_{i \in \mathcal{I}_{\text{forget}}}
a^{(\ell)}_n(\boldsymbol{\xi}_i).
\label{eq:mean-activations}
\end{equation}
Then, we define the corresponding directional bias
\vspace*{-0.4\baselineskip}
\begin{equation}
v^{(\ell)}_n
=
\mu^{(\ell)}_{\text{forget},n}
-
\mu^{(\ell)}_{\text{retain},n},
\label{eq:directional-bias}
\end{equation}
which quantifies the extent to which neuron $(\ell,n)$ is more (or less) activated by samples in $\mathcal{D}_{\text{forget}}$ than by those in $\mathcal{D}_{\text{retain}}$. A large magnitude of $v^{(\ell)}_n$ indicates that the neuron $(\ell,n)$ responds very differently to samples in the forget and retained subsets, meaning that its average activation is strongly shifted toward one group. Conversely, if the neuron exhibits similar activation patterns on both subsets, the two means nearly coincide and $v^{(\ell)}_n$ remains close to zero, indicating no preferential association. In this sense, $v^{(\ell)}_n$ serves as a simple directional statistic that measures how the neuron’s response changes between corrupted and reliable data,  in a similar spirit to neuron-level directional analyses used in concept activation studies~\cite{kim2018interpretability}. Neurons with large directional bias are therefore natural candidates for pruning, as their activations are disproportionately driven by patterns concentrated in the forget subset rather than by features that are physically meaningful across the dataset~\cite{ma2022learn}.

To account for differences in overall neuron activation levels, we further compute the empirical variance of neuron $(\ell,n)$ over the full dataset,
\vspace*{-0.25\baselineskip}
\begin{equation}
(\sigma^{(\ell)}_n)^2
=
\frac{1}{N}
\sum_{i=1}^N
\big(
a^{(\ell)}_n(\boldsymbol{\xi}_i)
-
\overline{a}^{(\ell)}_n
\big)^2,
\qquad
\overline{a}^{(\ell)}_n
=
\frac{1}{N}
\sum_{i=1}^N a^{(\ell)}_n(\boldsymbol{\xi}_i),
\label{eq:activation-variance}
\end{equation}
and define its bias-based importance score by
\begin{equation}
I^{(\ell)}_n
=
\frac{\big(v^{(\ell)}_n\big)^2}{(\sigma^{(\ell)}_n)^2 + \epsilon},
\label{eq:bias-importance}
\end{equation}
where $\epsilon > 0$ is a small regularization constant. The numerator in~\eqref{eq:bias-importance} penalizes large directional differences between forget and retained activations, while the denominator normalizes by the neuron's overall variability, thereby preventing highly active but unbiased neurons from being over-penalized. This ratio is analogous to the basic structure of an F-statistic \cite{davison2003statistical}, for which a between-group difference is scaled by a within-group variance. Here, however, we use a simpler two-group form tailored to activation differences rather than statistical testing. The score $I^{(\ell)}_n$ emphasizes neurons whose retain--forget shift is large compared with their overall activation variability, yielding a scale-normalized pruning criterion. Neurons are ranked by $I^{(\ell)}_n$ within each prunable layer, and those with the largest scores are removed first.


In practice, we may restrict pruning to a subset $\mathcal{L}_{\text{prune}} \subseteq \{1,\dots,L\}$ of layers (e.g., excluding the first and last hidden layers) to reduce the risk of excessively degrading the representation. For each prunable layer $\ell \in \mathcal{L}_{\text{prune}}$, we compute the importance scores $I^{(\ell)}_n$ defined in~\eqref{eq:bias-importance} and rank neurons accordingly. Then pruning is carried out from the most to the least biased according to a prescribed pruning schedule, as described next.

\subsection{Selective pruning and fine-tuning}
\label{subsec:pruning-ft}

Given the bias-based importance scores $\{I^{(\ell)}_n\}$ in~\eqref{eq:bias-importance}, we now describe the pruning and fine-tuning stage of P-PINN. Let $p \in (0,1)$ denote the total fraction of neurons to be pruned in each prunable layer, and let $K \in \mathbb{N}$ be the number of pruning iterations. We employ an iterative strategy in which,  at every iteration, a fraction $p/K$ of neurons is pruned from each prunable layer. After each pruning step, the importance scores are recomputed using the updated network. This process is repeated until the target pruning ratio $p$ is achieved. The complete pruning procedure is given in Algorithm~\ref{alg:ppinn} (see Part~II).

This iterative strategy has two advantages over single-shot pruning. First, recomputing importance scores after each pruning step mitigates the risk that pruning in one layer substantially changes the activation statistics in subsequent layers, which could invalidate the initial ranking. Second, distributing pruning over several iterations avoids overly aggressive one-shot removal of neurons and yields more stable performance. 
%

After completing all pruning iterations and reaching the target pruning ratio, we perform a fine-tuning stage of the pruned network using only the retained dataset $\mathcal{D}_{\text{retain}}$. In this stage, we optimize the same composite PINN loss~\eqref{eq:full-loss}, but with the data misfit term restricted to $\mathcal{D}_{\text{retain}}$:
\begin{equation}
\mathcal{J}_{\text{data}}^{\text{retain}}(\theta)
=
\frac{1}{|\mathcal{I}_{\text{retain}}|}
\sum_{i \in \mathcal{I}_{\text{retain}}}
\big\|
u_\theta(\boldsymbol{\xi}_i) - \bar{u}_i
\big\|_2^2.
\label{eq:data-loss-retain}
\end{equation}
The fine-tuning problem is thus
\begin{equation}
(\widetilde{\theta},\widetilde{\phi})
=
\arg\min_{\theta,\phi}
\Big\{
\lambda_{\text{data}} \, \mathcal{J}_{\text{data}}^{\text{retain}}(\theta)
+
\lambda_{\text{PDE}} \, \mathcal{J}_{\text{PDE}}(\theta,\phi)
+
\lambda_{\text{BC}} \, \mathcal{J}_{\text{BC}}(\theta,\phi)
+
\lambda_{\text{IC}} \, \mathcal{J}_{\text{IC}}(\theta,\phi)
\Big\},
\label{eq:ft-problem}
\end{equation}
with the optimization initialized using the parameters of the fully pruned network.

Using only $\mathcal{D}_{\text{retain}}$ in~\eqref{eq:data-loss-retain} encourages the pruned network to adapt its reduced representation to high-quality observations.
At the same time, the residual terms ensure compatibility with the PDE constraints over the full collocation sets. In practice, we employ a smaller learning rate and a limited number of fine-tuning epochs, as the optimization is warm-started from the noise-contaminated baseline model. The goal of this stage is not to retrain the network from scratch, but to refine the pruned architecture to improve predictive accuracy and robustness after the removal of noise-sensitive components.

\subsection{Summary and complexity analysis}
\label{subsec:algo-complexity}

\begin{algorithm}[t]
\caption{Selective pruning framework for noisy PINNs (P-PINN)}
\label{alg:ppinn}
\begin{algorithmic}[1]
\STATE \textbf{Input:} trained baseline PINN $(u_{\theta^0},\gamma_{\phi^0})$ obtained by minimizing~\eqref{eq:pinn-training} on the full dataset; noisy observations $\{(\boldsymbol{\xi}_i,\bar{u}_i)\}_{i=1}^N$; PDE, BC, IC collocation sets; loss weights $\lambda_{\text{data}},\lambda_{\text{PDE}},\lambda_{\text{BC}},\lambda_{\text{IC}}$; partition weights $\alpha_{\text{data}},\alpha_{\text{PDE}}$; retention ratio $\rho$; pruning ratio $p$; number of pruning iterations $K$; prunable layer set $\mathcal{L}_{\text{prune}}$.
\STATE \textbf{Output:} pruned and fine-tuned P-PINN $(u_{\widetilde{\theta}},\gamma_{\widetilde{\phi}})$.
\STATE \textbf{Part I: Data partitioning}
\STATE For each observation $i=1,\dots,N$, compute pointwise scores $r^{\text{data}}_i$ and $r^{\text{PDE}}_i$ via~\eqref{eq:point}, and form composite scores $M_i$ via~\eqref{eq:composite-metric}.
\STATE Choose threshold $\tau$ as the empirical $\rho$-quantile of $\{M_i\}$ and construct index sets $\mathcal{I}_{\text{retain}}$ and $\mathcal{I}_{\text{forget}}$ via~\eqref{eq:retain-forget-sets}, yielding datasets $\mathcal{D}_{\text{retain}}$ and $\mathcal{D}_{\text{forget}}$.
\STATE \textbf{Part II: Iterative selective pruning}
\FOR{$k = 1,\dots,K$}
  \STATE Evaluate the current network on $\{\boldsymbol{\xi}_i\}_{i=1}^N$ and compute the activations $a^{(\ell)}_n(\boldsymbol{\xi}_i)$ for all neurons in layers $\ell \in \mathcal{L}_{\text{prune}}$.
  \STATE Using the fixed partition $(\mathcal{I}_{\text{retain}},\mathcal{I}_{\text{forget}})$, compute the mean $\mu^{(\ell)}_{\text{retain},n}, \mu^{(\ell)}_{\text{forget},n}$, the directional biases $v^{(\ell)}_n$, and the importance scores $I^{(\ell)}_n$ via~\eqref{eq:mean-activations}--\eqref{eq:bias-importance}.
  \STATE For each prunable layer $\ell \in \mathcal{L}_{\text{prune}}$, identify the set $\mathcal{N}^{(\ell)}_k$ consisting of the neurons with the largest importance scores such that its cardinality $|\mathcal{N}^{(\ell)}_k|$ corresponds to pruning a fraction $p/K$ of the neurons that remain unpruned in layer $\ell$.
  \STATE Prune the neurons in $\mathcal{N}^{(\ell)}_k$, $\ell \in \mathcal{L}_{\text{prune}}$, by setting to zero the corresponding rows and columns in the weight matrices and the associated entries in the bias vectors of layers $\ell$ and $\ell+1$.

\ENDFOR
\STATE \textbf{Part III:  Selective fine-tuning on $\mathcal{D}_{\text{retain}}$}
\STATE Perform a fine-tuning of the pruned network on $\mathcal{D}_{\text{retain}}$ by solving~\eqref{eq:ft-problem}, yielding $(\widetilde{\theta},\widetilde{\phi})$.
\end{algorithmic}
\end{algorithm}

Algorithm~\ref{alg:ppinn} summarizes the complete P-PINN pipeline and fixes notation for the cost discussion below. We quantify the additional cost of partitioning and activation-based pruning relative to training the baseline PINN in~\eqref{eq:pinn-training}. 
From a computational perspective, baseline PINN training consists of $E_{\mathrm{train}}$ epochs, and each epoch is dominated by (i) a forward evaluation of the composite loss (data and physics residual terms) and (ii) a backward pass for gradient computation and the optimizer update. Let $C_{\mathrm{eval}}$ denote the cost of one such forward loss evaluation and let $C_{\mathrm{bwd}}$ denote the associated backward/update cost; then
\[
C_{\mathrm{train}}^{\mathrm{PINN}} \approx E_{\mathrm{train}}\,(C_{\mathrm{eval}}+C_{\mathrm{bwd}}).
\]
P-PINN adds: one extra evaluation pass to compute the partition scores (Part~I), $K$ additional evaluation passes to recompute activations and importance scores during iterative pruning (Part~II), and a short selective fine-tuning stage of $E_{\mathrm{ft}}$ epochs on the pruned network (Part~III), yielding the overhead
\[
C_{\mathrm{overhead}}^{\mathrm{P\text{-}PINN}} \approx (1+K)\,C_{\mathrm{eval}} + E_{\mathrm{ft}}\,(C_{\mathrm{eval}}+C_{\mathrm{bwd}}).
\]
Since $(1+K)$ is modest and $E_{\mathrm{ft}}\ll E_{\mathrm{train}}$ in all experiments, the additional cost is small relative to baseline training while providing substantial robustness gains.

\begin{figure}[t]
\centering
\begin{minipage}{0.32\textwidth}
  \centering
  \includegraphics[width=\textwidth]{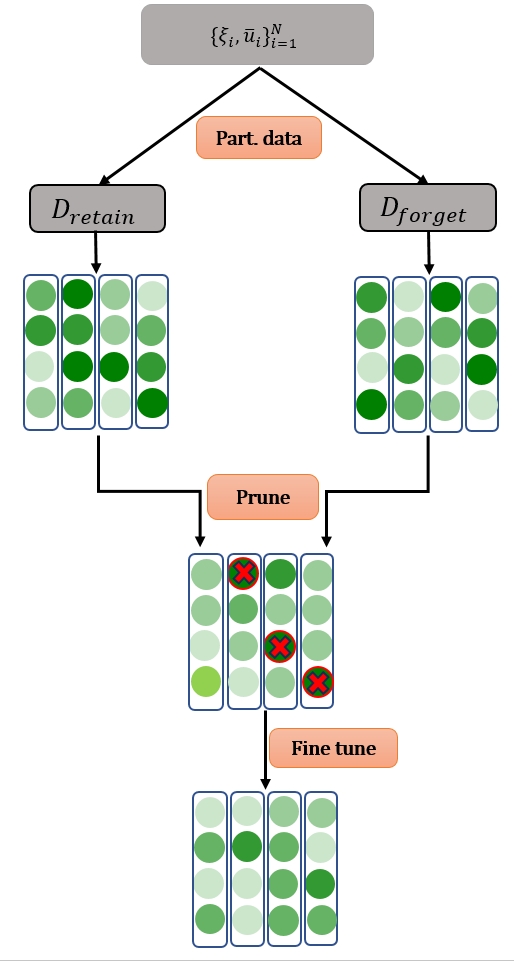}
\end{minipage}\hfill
\begin{minipage}{0.64\textwidth}
  \centering
  \includegraphics[width=\textwidth]{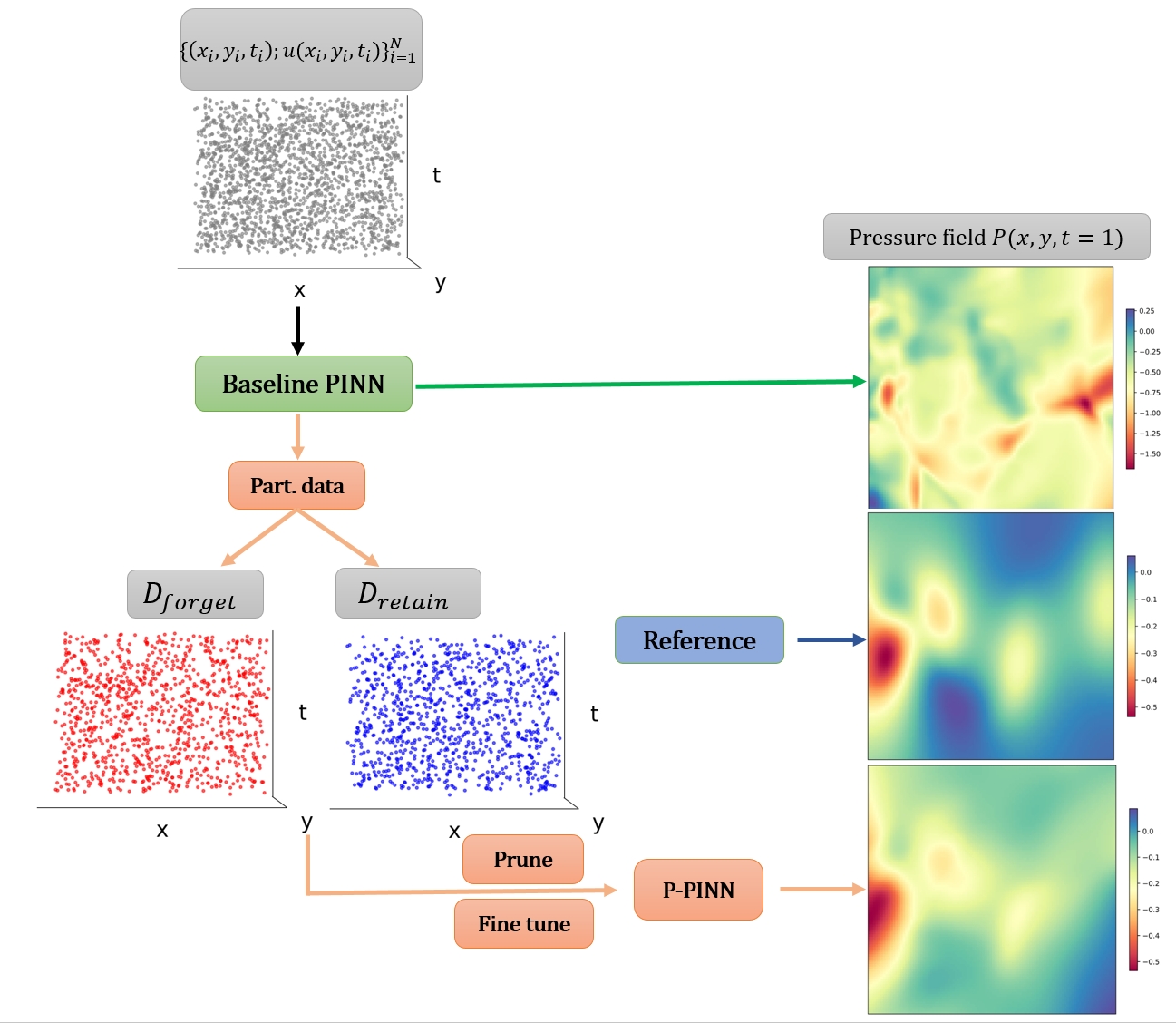}
\end{minipage}
\caption{\textbf{P-PINN workflow and a Navier--Stokes inverse example.}
\textbf{Left:} schematic of the P-PINN pipeline.
\textbf{Right:} Navier--Stokes inversion: A baseline PINN is trained on noisy space--time observations; the \emph{observations} are then partitioned into $\mathcal{D}_{\text{retain}}$ (blue) and $\mathcal{D}_{\text{forget}}$ (red) using residual--data-fidelity scores. The network is iteratively pruned and selectively fine-tuned on $\mathcal{D}_{\text{retain}}$ to obtain P-PINN. Color maps compare the baseline and P-PINN reconstructions of the pressure field $p(x,y,t{=}1)$ with the reference solution.}

\label{fig:ppinn-framework-ns}
\end{figure}

To illustrate the overall workflow of P-PINN and its effect on a representative case, Fig.~\ref{fig:ppinn-framework-ns} shows both an abstract schematic (left) and an NSInv instantiation (right). We first train a baseline PINN on the full set of noisy observations $\{(x_i,y_i,t_i;\bar u(x_i,y_i,t_i))\}_{i=1}^{N}$. Using the residual--data-fidelity scores, we then partition the observations into $\mathcal{D}_{\text{retain}}$ and $\mathcal{D}_{\text{forget}}$, prune neurons with the strongest activation bias toward $\mathcal{D}_{\text{forget}}$, and selectively fine-tune the pruned network on $\mathcal{D}_{\text{retain}}$. The color maps compare the baseline and P-PINN reconstructions of the pressure field $p(x,y,t{=}1)$ with the reference solution.


\section{Problem descriptions and numerical set-up}
\label{sec:numerics-setup}

We validate the proposed P-PINN framework on nine benchmark problems, comprising five parameter inversion tasks and four data assimilation tasks. This suite spans elliptic, parabolic, and hyperbolic PDEs, as well as the Navier--Stokes and Stokes equations, and is designed to thoroughly evaluate performance  under heterogeneous noise and partial observations. For each problem, we specify the governing PDE model, the domain configuration, the observation region, and the noise model.

\subsection{General experimental set-up}
\label{subsec:general-setup}

To rigorously evaluate robustness to corrupted data, we consider synthetic inverse problems with heterogeneous noise. In particular, for each problem, the exact solution $u^\star$ is sampled at $N$ locations, yielding perturbed observational data
\vspace*{-0.25\baselineskip}
\[
\bar{u}_i = u^\star(\boldsymbol{\xi}_i) + \eta_i,\qquad \eta_i \sim \mathcal{N}(0,\sigma_i^2).
\]
The noise standard deviations $\{\sigma_i\}$ are stratified into a high-noise level
$\sigma_{\mathrm{high}}$ and a low-noise level $\sigma_{\mathrm{low}}=0.1$.
Unless otherwise noted, a fraction of the observations (typically $60\%$) is corrupted with $\sigma_{\text{high}}$ to mimic severe sensor malfunction or outliers, while the remainder is perturbed with $\sigma_{\text{low}}$. For data assimilation tasks, the high/low split and $\sigma_{\text{high}}$ are adapted to the specific problem configuration (see Table~\ref{tab:benchmark_configs}).

Observation points are placed on uniform Cartesian grids in the relevant spatial or space--time domains, or in the prescribed observation subdomains for data assimilation. Collocation points for PDE residuals ($N_{\mathrm{PDE}}$) and boundary/initial conditions ($N_{\mathrm{BC}}, N_{\mathrm{IC}}$) are sampled on uniform grids for all parameter inversion problems. For data assimilation benchmarks, we instead employ Sobol low-discrepancy sequences~\cite{niederreiter1992random} for $N_{\mathrm{PDE}}$ to improve coverage in the space--time domain, while $N_{\mathrm{BC}}$ and $N_{\mathrm{IC}}$ remain on uniform grids. The dataset sizes, collocation counts, and noise configurations for all benchmarks are summarized in Table~\ref{tab:benchmark_configs}.

\begin{table}[htbp]
\centering
\caption{Summary of problem configurations. Here $N$ denotes the number of noisy observations, $N_{\mathrm{PDE}}$ the number of interior collocation points for enforcing the PDE residual, and $N_{\mathrm{BC/IC}}$ the numbers of boundary and initial collocation points (reported as $N_{\mathrm{BC}}/N_{\mathrm{IC}}$, with ``--'' indicating that a term is not present). The Noise column specifies the standard deviation for high/low corruption subsets and their respective proportions ($\sigma_{high}/\sigma_{low}$, Split \%).}
\label{tab:benchmark_configs}
\scalebox{0.8}{
\begin{tabular}{l c c c c}
\hline
\textbf{Problem} & $\mathbf{N}$ & $\mathbf{N_{PDE}}$ & $\mathbf{N_{BC/IC}}$ & \textbf{Noise} ($\sigma_{high}/\sigma_{low}$, Split) \\
\hline
\multicolumn{5}{l}{\textit{Parameter Inversion}} \\
PInv  & $2{,}500$ & $8{,}192$ & $2{,}048/-$ & $1.0/0.1$ ($60\%/40\%$) \\
HInv & $2{,}500$ & $8{,}192$ & $2{,}048/2{,}048$ & $1.0/0.1$ ($60\%/40\%$) \\
NSInv &  $10{,}000$ & $2{,}000$ & -- & $\sqrt{0.5}/0.1$ ($60\%/40\%$) \\
EBInv &  $10{,}000$ & $2{,}000$ & $100/200$ & $1.0/0.1$ ($60\%/40\%$) \\
WInv &  $10{,}000$ & $2{,}000$ & $100/200$ & $1.0/0.1$ ($60\%/40\%$) \\
\hline
\multicolumn{5}{l}{\textit{Data Assimilation}} \\
Poisson &  $225$ & $175$ & -- & $1.0/0.01$ ($20\%/80\%$) \\
Heat &  $400$ & $320$ & $80/-$ & $\sqrt{0.5}/0.1$ ($40\%/60\%$) \\
Wave &  $1{,}200$ & $2{,}160$ & $240/-$ & $\sqrt{0.5}/0.1$ ($20\%/80\%$) \\
Stokes & $320$ & $1{,}280$ & -- & $\sqrt{0.5}/0.1$ ($20\%/80\%$) \\
\hline
\end{tabular}}
\end{table}
\subsection{Parameter inversion problems}
\label{subsec:parameter-inversion}

In these benchmarks, the goal is to infer unknown coefficients (either scalar or spatially varying fields).

\subsubsection{Poisson coefficient inversion (PInv)}
\label{subsubsec:pinv}

We identify the spatially varying diffusivity $a^{*}(x)$ in the steady diffusion equation
\begin{equation}
\vspace*{-0.25\baselineskip}
-\nabla\cdot(a^{*}\nabla u) = f, \quad \mathbf{x} \in [0,1]^2,
\vspace*{-0.25\baselineskip}
\end{equation}
with exact solution $u^{*}(x,y)=\sin(\pi x)\sin(\pi y)$ and target coefficient
\vspace*{-0.25\baselineskip}
\begin{equation}
a^{*}(x,y) = \frac{1}{1+x^{2}+y^{2}+(x-1)^{2}+(y-1)^{2}}.
\end{equation}
Data, collocation sets, and noise levels follow Table~\ref{tab:benchmark_configs}; the task is to reconstruct $a^{*}$ from noisy solution measurements. For evaluation, we use an independent noise-free test set of $10{,}000$ samples on a $100 \times 100$ grid.

\subsubsection{Heat (diffusion) inversion (HInv)}
\label{subsubsec:hinv}

We reconstruct the coefficient $a^{*}(x,y)$ in the time-dependent heat equation
\begin{equation}
\partial_{t}u - \nabla\cdot(a^{*}\nabla u) = f, \quad (\mathbf{x},t) \in [-1,1]^2 \times [0,1],
\end{equation}
with exact fields $u^{*}(x,y,t)=e^{-t}\sin(\pi x)\sin(\pi y)$ and $a^{*}(x,y)=2+\sin(\pi x)\sin(\pi y)$. Noisy space--time observations and collocation sets are configured as in Table~\ref{tab:benchmark_configs}, and the test set consists of $180{,}000$ noise-free samples on a $60 \times 60 \times 50$ Cartesian grid.

\subsubsection{Navier--Stokes inversion (NSInv)}
\label{subsubsec:nsinv}

We consider the 2D incompressible Navier--Stokes equations for flow past a cylinder on
$\Omega \times I = [1,8]\times[-2,2]\times[0,7]$,
\vspace*{-0.4\baselineskip}
\begin{equation}
\partial_{t}\mathbf{u} + \beta_1 (\mathbf{u}\cdot\nabla)\mathbf{u}
= -\nabla p + \beta_{2}\Delta \mathbf{u}, \quad \nabla\cdot \mathbf{u} = 0.
\end{equation}
Following~\cite{raissi2019physics}, we use their cylinder-flow dataset and partition the available interior velocity
snapshots into an observational set used for training (to which we add heterogeneous Gaussian noise to form the
perturbed observations) and a held-out set used only for testing. The true parameters are $\beta_{1}=1$ and $\beta_{2}=0.01$.
Noisy interior velocity samples and collocation sets follow Table~\ref{tab:benchmark_configs}.
We infer $(\beta_{1},\beta_{2})$ and the pressure field $p$ from the training velocity data, and evaluate on the held-out velocity samples.

\subsubsection{Euler--Bernoulli beam inversion (EBInv)}
\label{subsubsec:ebinv}

We estimate the parameter $\alpha$ in the beam equation
\vspace*{-0.4\baselineskip}
\begin{equation}
\partial_{tt}u + \alpha^{2}\partial_{xxxx}u = 0, \quad (x,t) \in [0,1]^2,
\end{equation}
with analytical solution $u^{*}(x,t)=\sin(\pi x)\cos(\pi^{2}t)$ for $\alpha=1$. Noisy samples on a uniform space--time grid and collocation sets are chosen according to Table~\ref{tab:benchmark_configs}; the inverse task is to recover $\alpha$ (or $\alpha^{2}$). 

\subsubsection{Wave inversion (WInv)}
\label{subsubsec:winv}

We infer the wave speed $c$ in
\begin{equation}
\partial_{tt}u - c^{2}\partial_{xx}u = 0, \quad (x,t) \in [0,1]^2,
\end{equation}
with exact solution for $c=2$,
\begin{equation}
u^{*}(x,t)
=
\sin(\pi x)\cos(c\pi t)
+
0.5\sin(4\pi x)\cos(4c\pi t).
\end{equation}
Noisy observations and collocation sets follow Table~\ref{tab:benchmark_configs}, and we reconstruct $c$ (or $c^{2}$). 


\subsection{Data assimilation problems}
\label{subsec:da-problems}

In the data assimilation benchmarks, the PDE parameters are known, but data are restricted to a partial subdomain $\Omega^{\prime}\subset\Omega$. The goal is to reconstruct the full solution from partial, noisy observations.

\subsubsection{Poisson data assimilation (Poisson)}

We consider the Poisson problem
\begin{equation}
-\Delta u = f \quad \text{in } \Omega=(0,1)^2,
\end{equation}
with exact solution $u^{*}(x_{1},x_{2})=30x_{1}x_{2}(1-x_{1})(1-x_{2})$. Observations are restricted to the central region
\[
\Omega^{\prime}=\big\{(x_{1},x_{2})\in\Omega:|x_{1}-0.5|< 0.375,\ |x_{2}-0.5|<0.375\big\},
\]
and corrupted by heterogeneous Gaussian noise as in Table~\ref{tab:benchmark_configs}. PDE residuals are enforced at interior collocation points, and reconstructions are evaluated on a $200\times 200$ noise-free grid over $\Omega$.

\subsubsection{Heat data assimilation (Heat)}

We consider
\begin{equation}
\partial_{t}u - \Delta u = f
\quad \text{in } \Omega_T = (0,1)\times(0,0.02),
\end{equation}
with exact solution $u^{*}(x,t)=\sin(2\pi x)e^{-4\pi^{2}t}$. Data are available only in the spatial band
\[
\Omega^{\prime}=(0.2,0.8), \qquad \Omega^{\prime}_T = \Omega^{\prime}\times(0,T),
\]
and are corrupted as specified in Table~\ref{tab:benchmark_configs}. Accuracy is measured on a $200\times 200$ noise-free grid in $\Omega_T$.

\subsubsection{Wave data assimilation (Wave)}

We consider
\begin{equation}
\partial_{tt}u - c^{2}\partial_{xx}u = 0
\quad \text{in } \Omega_T = (0,1)\times(0,1),
\end{equation}
with $c=1$ and exact solution $u^{*}(x,t)=\sin(2\pi x)\sin(2\pi t)$. Observations are restricted to
\[
\Omega^{\prime} = (0,0.2)\cup(0.8,1),\qquad \Omega^{\prime}_T = \Omega^{\prime}\times(0,1),
\]
with heterogeneous noise as in Table~\ref{tab:benchmark_configs}. Evaluation is performed on a $200 \times 200$ noise-free grid.

\subsubsection{Stokes data assimilation (Stokes)}

We reconstruct velocity and pressure for stationary Stokes flow on $\Omega=(0,1)^2$,
\begin{equation}
\Delta \mathbf{u} + \nabla p = \mathbf{f},\qquad
\nabla\cdot\mathbf{u} = f_d,
\end{equation}
with homogeneous forcing and exact fields
\[
\mathbf{u}^\star(x_1,x_2) = \big(4x_1 x_2^3,\; x_1^4 - x_2^4\big),
\qquad
p^\star(x_1,x_2) = 12 x_1^2 x_2 - 4x_2^3 - 1.
\]
Velocity observations are available only in the circular subdomain
\[
\Omega^{\prime}=\big\{(x_{1},x_{2})\in\Omega:\|(x_{1},x_{2})-(0.5,0.5)\|_{2}<0.25\big\},
\]
with noise levels and sampling as in Table~\ref{tab:benchmark_configs}. PDE residuals are enforced in $\Omega$, and evaluation uses a $200\times 200$ noise-free grid for $(\mathbf{u}^\star,p^\star)$.

\subsection{Network architecture and training setup}
\label{subsec:arch-training}

Unless otherwise stated, all experiments use a fully connected MLP with five hidden layers of width $100$ and $\tanh$ activation functions. The network maps physical coordinates (and time, when applicable) to the state $u_\theta$ and, in inverse problems, to the parameter field $\gamma_\phi$. For spatially varying parameters, $u_\theta$ and $\gamma_\phi$ share a common backbone and are produced by two linear heads; for global scalar coefficients, the entries of $\gamma$ are modeled as free trainable variables. This yields a uniform architecture across benchmarks while allowing joint learning of states and parameters.

For each problem, we perform three independent runs with random seeds $42$, $43$, and $44$. In parameter inversion problems, the baseline PINN is trained for $20{,}000$ epochs using the Adam optimizer with learning rate $10^{-3}$ and momentum parameters $\beta_1^{\text{Adam}}=0.9$, $\beta_2^{\text{Adam}}=0.999$. For data assimilation problems, we follow the data-guided PINNs (DG-PINN) strategy~\cite{zhou2024data}: a \emph{pre-training} phase with $20{,}000$ Adam epochs using only the data loss, followed by a \emph{fine-tuning} phase of $5{,}000$ L-BFGS iterations minimizing the full composite loss, with history size $50$ and gradient-norm tolerance $10^{-8}$. All experiments are performed on a single NVIDIA A100 GPU with memory usage below $1$~GB.

Loss weights in the baseline training are set to unity,
\(
\lambda_{\mathrm{PDE}} = \lambda_{\mathrm{BC}} = \lambda_{\mathrm{IC}} = \lambda_{\mathrm{data}} = 1,
\)
for both inverse and data assimilation problems. After selective pruning, we reduce the PDE-residual weight to
\(
\lambda_{\mathrm{PDE}} = 0.005
\)
for fine-tuning on $\mathcal{D}_{\text{retain}}$, while keeping the other weights unchanged. This avoids over-emphasizing the physics term once the effective dataset has been reduced.

\subsection{Data partitioning and pruning schedule}
\label{subsec:partition-pruning-detail}

For each trained baseline PINN, we compute the per-observation composite scores $M_i$ defined in Section~\ref{subsec:partition}. In practice, we approximate the metric~\eqref{eq:composite-metric} with $\alpha_{\text{data}}=1$ and $\alpha_{\text{PDE}}=0.001$. The weights were chosen so that the data misfit and PDE residual have comparable magnitudes, while still allowing the data term to dominate in regions where the PDE residual is small. A threshold $\tau$ is then chosen, initially from the grid $\{0.01,0.1,1\}$ and subsequently adjusted based on the empirical distribution of $\{M_i\}$, to obtain a retention ratio $\rho = |\mathcal{D}_{\text{retain}}|/N$ in a moderate range (typically $40\%$–$80\%$). This yields a partition $(\mathcal{D}_{\text{retain}},\mathcal{D}_{\text{forget}})$ as in~\eqref{eq:retain-forget-sets}.

Neuron importance scores are computed using the bias-based metric introduced in Section~\ref{subsec:bias-score}. We adopt an iterative pruning schedule applied to the first five hidden layers (excluding the output layer). Each pruning iteration removes $5\%$ of the neurons in the selected layers, based on the current importance scores. We perform $20$ pruning iterations, for a total pruning ratio of approximately $64\%$ in the prunable layers; the pruning ratio and the number of iterations were selected to strike a balance between compactness and accuracy. Final fine-tuning after the pruning stage uses $2{,}000$ Adam epochs for parameter inversion problems, and $2{,}000$ Adam epochs followed by $500$ L-BFGS iterations for data assimilation problems. The additional training cost is roughly one-tenth of the initial baseline training cost. We compare this iterative scheme with single-step pruning and with alternative importance metrics in Section~\ref{subsec:training-efficiency}.
\section{Numerical results}
\label{sec:numerical-results}

We now present numerical results for the nine benchmark problems described in Section~\ref{sec:numerics-setup}. After briefly recalling the evaluation metrics and baseline strategies, we first report aggregate accuracy across all tasks. We then provide qualitative case studies that visualize representative reconstructions. Next, we examine training efficiency together with the impact of different pruning strategies and neuron-importance criteria. Finally, we conduct sensitivity analyses with respect to the observational noise level, pruning depth, and data-retention ratio.

\subsection{Evaluation metrics and baseline methods}
\label{subsec:metrics-baselines}

Let $\hat{\mathbf{q}}=(\hat{q}_1,\dots,\hat{q}_n)$ and $\mathbf{q}=(q_1,\dots,q_n)$ denote the predicted solutions and the ground truth/reference (e.g., state $u^\star$ or parameter field $\gamma^\star$). We report the relative $L^2$ error
\vspace*{-0.25\baselineskip}
\begin{equation}
\mathrm{L2RE}(\mathbf{q})
=
\frac{\|\hat{\mathbf{q}}-\mathbf{q}\|_2}{\|\mathbf{q}\|_2}
=
\sqrt{\frac{\sum_{i=1}^n (\hat{q}_i-q_i)^2}{\sum_{i=1}^n q_i^2}},
\label{eq:l2re-def}
\end{equation}
along with the $L^1$ relative error (L1RE), mean squared error (MSE), and maximum absolute error (MAE),

\begin{equation}
\mathrm{L1RE}(\mathbf{q})=
\frac{\sum_{i=1}^n |\hat{q}_i-q_i|}{\sum_{i=1}^n |q_i|},
\mathrm{MSE}(\mathbf{q})=
\frac{1}{n}\sum_{i=1}^n (\hat{q}_i-q_i)^2,
\label{eq:l1re-mse-def}
\mathrm{MAE}(\mathbf{q})
=\max_{1\le i\le n}|\hat{q}_i-q_i|.
\end{equation}
We also report Fourier-band mean squared errors (fMSE) in low-, mid-, and high-frequency ranges by restricting the discrete Fourier transforms of
$\mathbf{q}$ and $\hat{\mathbf{q}}$ to a band $[k_{\min},k_{\max}]$ following~\cite{takamoto2022pdebench}, denoted by
$\mathrm{fMSE}_\mathrm{L}$, $\mathrm{fMSE}_\mathrm{M}$, and $\mathrm{fMSE}_\mathrm{H}$.

We compare the proposed P-PINN with three training pipelines:
\begin{itemize}
\item \emph{Baseline PINN (PINN)}: the standard PINN described in Section~\ref{subsec:pinn-formulation}, trained on the full noisy dataset.
\item \emph{Fine-tune on $\mathcal{D}_{\text{retain}}$ (FT)}: the baseline PINN further trained on the retained dataset $\mathcal{D}_{\text{retain}}$, but without pruning.
\item \emph{Retrain-on-$\mathcal{D}_{\text{retain}}$ (RT)}: a new PINN trained from scratch on $\mathcal{D}_{\text{retain}}$ and the same collocation sets, without reusing the baseline weights.
\end{itemize}
We refer to these methods as PINN, FT, and RT, respectively, throughout this section. In addition, we consider two uncertainty-aware baselines with the same network architecture and composite loss as PINN. 
\emph{Dropout-PINN} (D-PINN) \cite{zhang2019quantifying} inserts dropout with rate $0.1$ after each hidden layer and uses Monte Carlo dropout with $50$ stochastic forward passes at inference. 
\emph{B-PINN} \cite{yang2021b} imposes an i.i.d.\ zero-mean Gaussian prior with unit variance on all network weights, initializes Hamiltonian Monte Carlo from the PINN solution, and collects $50$ posterior samples (after $20$ burn-in steps), using the posterior predictive mean for evaluation. For a fair comparison, the uncertainty-aware baselines B-PINN and D-PINN inherit exactly the same training protocol as the corresponding baseline PINN.

For ablations we also compare single-step pruning and several classical activation-based pruning criteria, while keeping the overall pipeline and hyperparameters fixed. Unless otherwise stated, all reported numbers correspond to the mean and standard deviation over three independent runs. Global accuracy tables focus on PINN, P-PINN, B-PINN, and D-PINN, while FT and RT are primarily used in Section~\ref{subsec:training-efficiency} to analyze computational trade-offs.

\subsection{Global accuracy and spectral error analysis}
\label{subsec:main-res-sisc}
\label{subsec:other-metrics}

We first assess the reconstruction accuracy of the four PINN variants on all
nine benchmarks. Table~\ref{tab:L2RE} reports the mean and standard deviation
of the L2RE for the four data assimilation problems and the five parameter inversion
problems, comparing the baseline PINN, the proposed P-PINN, the B-PINN, and the D-PINN.

\begin{table}[!htbp]
\centering
\caption{Mean $\pm$ std of L2RE for data assimilation and parameter inversion.}
\label{tab:L2RE}
\begingroup
\setlength{\tabcolsep}{2pt}
\footnotesize
\scalebox{0.85}{\begin{tabular}{c|cccc|ccccc}
\toprule
\textbf{L2RE} & \multicolumn{4}{c|}{\textbf{Data assimilation}} & \multicolumn{5}{c}{\textbf{Parameter inversion}} \\
\midrule
              & Poisson & Heat    & Wave     & Stokes   & PInv     & HInv     & NSInv    & EBInv    & WInv      \\
\midrule
\multirow{2}{*}{PINN}
  & $2.59\mathrm{E}{-}1$ & $9.63\mathrm{E}{-}1$ & $7.66\mathrm{E}{-}1$ & $6.68\mathrm{E}{-}1$ 
  & $9.37\mathrm{E}{-}2$ & $1.32\mathrm{E}{-}1$ & $3.44\mathrm{E}{+}0$ & $2.50\mathrm{E}{-}1$ & $9.94\mathrm{E}{-}2$ \\
  & {\scriptsize$\pm1.34\mathrm{E}{-}2$} 
  & {\scriptsize$\pm2.19\mathrm{E}{-}1$} 
  & {\scriptsize$\pm6.81\mathrm{E}{-}2$} 
  & {\scriptsize$\pm6.50\mathrm{E}{-}3$} 
  & {\scriptsize$\pm2.44\mathrm{E}{-}2$} 
  & {\scriptsize$\pm5.78\mathrm{E}{-}3$} 
  & {\scriptsize$\pm1.09\mathrm{E}{+}0$} 
  & {\scriptsize$\pm1.96\mathrm{E}{-}2$} 
  & {\scriptsize$\pm2.60\mathrm{E}{-}3$} \\
\midrule
\multirow{2}{*}{P-PINN}
  & $\mathbf{8.04\mathrm{E}{-}2}$ & $\mathbf{3.26\mathrm{E}{-}2}$ & $\mathbf{8.88\mathrm{E}{-}2}$ & $\mathbf{1.19\mathrm{E}{-}1}$ 
  & $\mathbf{1.82\mathrm{E}{-}2}$ & $\mathbf{8.14\mathrm{E}{-}2}$ & $6.44\mathrm{E}{-}1$ & $\mathbf{6.95\mathrm{E}{-}2}$ & $\mathbf{5.71\mathrm{E}{-}2}$ \\
& {\scriptsize$\mathbf{\pm3.13\mathrm{E}{-}2}$} 
  & {\scriptsize$\mathbf{\pm1.56\mathrm{E}{-}4}$} 
  & {\scriptsize$\mathbf{\pm4.38\mathrm{E}{-}4}$} 
  & {\scriptsize$\mathbf{\pm2.03\mathrm{E}{-}2}$} 
  & {\scriptsize$\mathbf{\pm2.17\mathrm{E}{-}3}$} 
  & {\scriptsize$\mathbf{\pm3.86\mathrm{E}{-}3}$} 
  & {\scriptsize$\pm2.67\mathrm{E}{-}1$} 
  & {\scriptsize$\mathbf{\pm7.14\mathrm{E}{-}3}$} 
  & {\scriptsize$\mathbf{\pm3.46\mathrm{E}{-}3}$} \\
\midrule
\multirow{2}{*}{B-PINN}
  & $3.67\mathrm{E}{-}1$ & $3.94\mathrm{E}{-}1$ & $6.97\mathrm{E}{-}1$ & $6.95\mathrm{E}{-}1$ 
  & $5.70\mathrm{E}{-}1$ & $2.68\mathrm{E}{-}1$ & $1.98\mathrm{E}{+}0$ & $7.82\mathrm{E}{-}1$ & $2.12\mathrm{E}{-}1$ \\
  & {\scriptsize$\pm1.14\mathrm{E}{-}1$} 
  & {\scriptsize$\pm1.08\mathrm{E}{-}1$} 
  & {\scriptsize$\pm1.26\mathrm{E}{-}1$} 
  & {\scriptsize$\pm1.21\mathrm{E}{-}1$} 
  & {\scriptsize$\pm9.39\mathrm{E}{-}2$} 
  & {\scriptsize$\pm5.62\mathrm{E}{-}2$} 
  & {\scriptsize$\pm1.84\mathrm{E}{-}1$} 
  & {\scriptsize$\pm1.98\mathrm{E}{-}2$} 
  & {\scriptsize$\pm1.51\mathrm{E}{-}1$} \\
\midrule
\multirow{2}{*}{D-PINN}
  & $2.17\mathrm{E}{-}1$ & $3.16\mathrm{E}{-}1$ & $1.35\mathrm{E}{-}1$ & $5.46\mathrm{E}{-}1$ 
  & $1.01\mathrm{E}{-}1$ & $2.21\mathrm{E}{-}1$ & $\mathbf{2.28\mathrm{E}{-}1}$ & $3.90\mathrm{E}{-}1$ & $1.57\mathrm{E}{-}1$ \\
  & {\scriptsize$\pm2.78\mathrm{E}{-}3$} 
  & {\scriptsize$\pm1.57\mathrm{E}{-}2$} 
  & {\scriptsize$\pm1.08\mathrm{E}{-}2$} 
  & {\scriptsize$\pm8.49\mathrm{E}{-}3$} 
  & {\scriptsize$\pm1.28\mathrm{E}{-}2$} 
  & {\scriptsize$\pm2.21\mathrm{E}{-}3$} 
  & {\scriptsize$\mathbf{\pm1.15\mathrm{E}{-}2}$} 
  & {\scriptsize$\pm2.83\mathrm{E}{-}1$} 
  & {\scriptsize$\pm7.28\mathrm{E}{-}2$} \\
\bottomrule
\end{tabular}}
\endgroup
\end{table}

Table~\ref{tab:L2RE} demonstrates that P-PINN consistently improves L2RE over the baseline PINN on all nine
benchmarks and attains the lowest L2RE on seven of them. The improvements are most
pronounced on the heat and wave data assimilation problems under strong noise:
for the heat equation, P-PINN reduces L2RE from $9.63\times 10^{-1}$ to
$3.26\times 10^{-2}$, while for the
wave equation L2RE decreases from $7.66\times 10^{-1}$ to $8.88\times 10^{-2}$.
D-PINN attains the smallest L2RE on NSInv, but it exhibits substantially
larger errors than P-PINN on most of the remaining problems and therefore provides
less uniform performance across the full suite of benchmarks.

To verify that these trends are not specific to the $\mathrm{L}^2$ norm,
Tables~\ref{tab:L1RE} and~\ref{tab:MSE} report the mean and standard
deviation of the L1RE and MSE, respectively. Across most benchmarks, and in particular across the parameter inversion tasks, P-PINN yields the smallest L1RE and MSE,
typically improving upon the baseline PINN by factors of $3$--$10$. D-PINN
provides very competitive errors on NSInv (and in particular the best state
MSE on that problem), but at the cost of noticeably degraded accuracy on
other inverse problems such as EBInv and WInv. In contrast, P-PINN delivers
consistent gains over the baseline PINN and compares favorably with both
B-PINN and D-PINN on the majority of benchmarks.

\begin{table}[htbp]
\centering
\caption{Mean $\pm$ std of L1RE.}
\scalebox{0.8}{
\setlength{\tabcolsep}{2pt}
\begin{tabular}{c|cccc|ccccc}
\toprule
\textbf{L1RE} & \multicolumn{4}{c|}{\textbf{Data assimilation}} & \multicolumn{5}{c}{\textbf{Parameter inversion}} \\
\midrule
              & Poisson & Heat & Wave & Stokes & PInv & HInv & NSInv & EBInv & WInv \\
\midrule
\multirow{2}{*}{PINN}
  & $1.85\mathrm{E}{-}1$ & $5.34\mathrm{E}{-}1$ & $1.45\mathrm{E}{-}1$ & $6.04\mathrm{E}{-}1$ 
  & $6.44\mathrm{E}{-}2$ & $1.03\mathrm{E}{-}1$ & $3.80\mathrm{E}{+}0$ & $2.50\mathrm{E}{-}1$ & $9.94\mathrm{E}{-}2$ \\
  & {\scriptsize$\pm6.13\mathrm{E}{-}3$} 
  & {\scriptsize$\pm2.17\mathrm{E}{-}1$} 
  & {\scriptsize$\pm5.50\mathrm{E}{-}2$} 
  & {\scriptsize$\pm7.30\mathrm{E}{-}3$} 
  & {\scriptsize$\pm1.97\mathrm{E}{-}2$} 
  & {\scriptsize$\pm3.06\mathrm{E}{-}3$} 
  & {\scriptsize$\pm8.43\mathrm{E}{-}1$} 
  & {\scriptsize$\pm1.96\mathrm{E}{-}2$} 
  & {\scriptsize$\pm2.60\mathrm{E}{-}3$} \\
\midrule
\multirow{2}{*}{P-PINN}
 & $\mathbf{6.04\mathrm{E}{-}2}$ & $\mathbf{1.35\mathrm{E}{-}2}$ & $\mathbf{2.50\mathrm{E}{-}2}$ & $\mathbf{1.17\mathrm{E}{-}1}$ 
& $\mathbf{1.94\mathrm{E}{-}2}$ & $\mathbf{6.20\mathrm{E}{-}2}$ & $1.49\mathrm{E}{+}0$ & $\mathbf{6.95\mathrm{E}{-}2}$ & $\mathbf{5.72\mathrm{E}{-}2}$ \\
& {\scriptsize$\mathbf{\pm2.23\mathrm{E}{-}2}$} 
& {\scriptsize$\mathbf{\pm6.63\mathrm{E}{-}3}$} 
& {\scriptsize$\mathbf{\pm2.08\mathrm{E}{-}3}$} 
& {\scriptsize$\mathbf{\pm1.49\mathrm{E}{-}2}$} 
& {\scriptsize$\mathbf{\pm2.26\mathrm{E}{-}3}$} 
& {\scriptsize$\mathbf{\pm3.01\mathrm{E}{-}3}$} 
& {\scriptsize$\pm5.30\mathrm{E}{-}1$} 
& {\scriptsize$\mathbf{\pm7.14\mathrm{E}{-}3}$} 
& {\scriptsize$\mathbf{\pm3.46\mathrm{E}{-}3}$} \\
\midrule
\multirow{2}{*}{B-PINN}
  & $3.37\mathrm{E}{-}1$ & $4.06\mathrm{E}{-}1$ & $7.16\mathrm{E}{-1}$ & $9.95\mathrm{E}{-}1$ 
  & $4.67\mathrm{E}{-}1$ & $2.24\mathrm{E}{-}1$ & $2.26\mathrm{E}{+}0$ & $7.82\mathrm{E}{-}1$ & $2.12\mathrm{E}{-}1$ \\
  & {\scriptsize$\pm1.14\mathrm{E}{-}1$} 
  & {\scriptsize$\pm1.13\mathrm{E}{-}2$} 
  & {\scriptsize$\pm1.44\mathrm{E}{-}1$} 
  & {\scriptsize$\pm2.51\mathrm{E}{-}1$} 
  & {\scriptsize$\pm8.70\mathrm{E}{-}2$} 
  & {\scriptsize$\pm4.16\mathrm{E}{-}2$} 
  & {\scriptsize$\pm1.98\mathrm{E}{-}1$} 
  & {\scriptsize$\pm1.98\mathrm{E}{-}2$} 
  & {\scriptsize$\pm1.51\mathrm{E}{-}1$} \\
\midrule
\multirow{2}{*}{D-PINN}
  & $1.88\mathrm{E}{-}1$ & $3.02\mathrm{E}{-}1$ & $1.34\mathrm{E}{-}1$ & $5.88\mathrm{E}{-}1$ 
  & $7.52\mathrm{E}{-}2$ & $1.82\mathrm{E}{-}1$ & $\mathbf{2.27\mathrm{E}{-}1}$ & $3.90\mathrm{E}{-}1$ & $1.57\mathrm{E}{-}1$ \\
  & {\scriptsize$\pm1.57\mathrm{E}{-}2$} 
  & {\scriptsize$\pm1.49\mathrm{E}{-}2$} 
  & {\scriptsize$\pm1.05\mathrm{E}{-}2$} 
  & {\scriptsize$\pm1.06\mathrm{E}{-}2$} 
  & {\scriptsize$\pm1.09\mathrm{E}{-}2$} 
  & {\scriptsize$\pm2.06\mathrm{E}{-}3$} 
  & {\scriptsize$\mathbf{\pm1.61\mathrm{E}{-}2}$} 
  & {\scriptsize$\pm2.83\mathrm{E}{-}1$} 
  & {\scriptsize$\pm7.28\mathrm{E}{-}2$} \\
\bottomrule
\end{tabular}}
\label{tab:L1RE}
\end{table}

\begin{table}[htbp]
\centering
\caption{Mean $\pm$ std of MSE.}
\scalebox{0.8}{
\setlength{\tabcolsep}{2pt}
\begin{tabular}{c|cccc|ccccc}
\toprule
\textbf{MSE} & \multicolumn{4}{c|}{\textbf{Data assimilation}} & \multicolumn{5}{c}{\textbf{Parameter inversion}} \\
\midrule
              & Poisson & Heat & Wave & Stokes & PInv & HInv & NSInv & EBInv & WInv \\
\midrule
\multirow{2}{*}{PINN}
  & $6.68\mathrm{E}{-}2$ & $2.00\mathrm{E}{-}1$ & $7.54\mathrm{E}{-}3$ & $2.08\mathrm{E}{-}1$ 
  & $6.24\mathrm{E}{-}2$ & $7.44\mathrm{E}{-}2$ & $2.73\mathrm{E}{-}1$ & $6.22\mathrm{E}{-}2$ & $3.95\mathrm{E}{-}2$ \\
  & {\scriptsize$\pm7.14\mathrm{E}{-}3$} 
  & {\scriptsize$\pm1.42\mathrm{E}{-}1$} 
  & {\scriptsize$\pm5.87\mathrm{E}{-}3$} 
  & {\scriptsize$\pm6.80\mathrm{E}{-}3$} 
  & {\scriptsize$\pm1.90\mathrm{E}{-}2$} 
  & {\scriptsize$\pm2.62\mathrm{E}{-}3$} 
  & {\scriptsize$\pm1.47\mathrm{E}{-}1$} 
  & {\scriptsize$\pm9.75\mathrm{E}{-}3$} 
  & {\scriptsize$\pm2.03\mathrm{E}{-}3$} \\
\midrule
\multirow{2}{*}{P-PINN}
  & $\mathbf{7.50\mathrm{E}{-}3}$ & $\mathbf{6.43\mathrm{E}{-}4}$ & $\mathbf{1.57\mathrm{E}{-}4}$ & $\mathbf{6.60\mathrm{E}{-}3}$ 
& $\mathbf{1.32\mathrm{E}{-}4}$ & $\mathbf{2.82\mathrm{E}{-}2}$ & $2.48\mathrm{E}{-}2$ & $\mathbf{4.84\mathrm{E}{-}3}$ & $\mathbf{1.29\mathrm{E}{-}2}$ \\
& {\scriptsize$\mathbf{\pm5.73\mathrm{E}{-}3}$} 
& {\scriptsize$\mathbf{\pm9.67\mathrm{E}{-}3}$} 
& {\scriptsize$\mathbf{\pm2.89\mathrm{E}{-}5}$} 
& {\scriptsize$\mathbf{\pm2.10\mathrm{E}{-}3}$} 
& {\scriptsize$\mathbf{\pm3.12\mathrm{E}{-}5}$} 
& {\scriptsize$\mathbf{\pm5.02\mathrm{E}{-}2}$} 
& {\scriptsize$\pm1.16\mathrm{E}{-}2$} 
& {\scriptsize$\mathbf{\pm9.92\mathrm{E}{-}4}$} 
& {\scriptsize$\mathbf{\pm1.58\mathrm{E}{-}3}$} \\ 
\midrule
\multirow{2}{*}{B-PINN}
  & $1.46\mathrm{E}{-}1$ & $8.26\mathrm{E}{-}2$ & $1.24\mathrm{E}{-}1$ & $2.29\mathrm{E}{-}1$ 
  & $1.32\mathrm{E}{-}4$ & $3.19\mathrm{E}{-}1$ & $1.21\mathrm{E}{+}0$ & $6.13\mathrm{E}{-}1$ & $2.78\mathrm{E}{-}2$ \\
  & {\scriptsize$\pm8.19\mathrm{E}{-}2$} 
  & {\scriptsize$\pm3.45\mathrm{E}{-}4$} 
  & {\scriptsize$\pm4.14\mathrm{E}{-}2$} 
  & {\scriptsize$\pm7.27\mathrm{E}{-}2$} 
  & {\scriptsize$\pm3.12\mathrm{E}{-}5$} 
  & {\scriptsize$\pm1.28\mathrm{E}{-}1$} 
  & {\scriptsize$\pm2.26\mathrm{E}{-}1$} 
  & {\scriptsize$\pm3.12\mathrm{E}{-}2$} 
  & {\scriptsize$\pm2.78\mathrm{E}{-}3$} \\
\midrule
\multirow{2}{*}{D-PINN}
  & $4.65\mathrm{E}{-}2$ & $4.93\mathrm{E}{-}2$ & $4.54\mathrm{E}{-}3$ & $1.37\mathrm{E}{-}1$ 
  & $1.95\mathrm{E}{-}3$ & $2.08\mathrm{E}{-}1$ & $\mathbf{1.51\mathrm{E}{-}2}$ & $2.32\mathrm{E}{-}1$ & $2.23\mathrm{E}{-}2$ \\
  & {\scriptsize$\pm1.19\mathrm{E}{-}3$} 
  & {\scriptsize$\pm4.84\mathrm{E}{-}3$} 
  & {\scriptsize$\pm7.28\mathrm{E}{-}4$} 
  & {\scriptsize$\pm4.28\mathrm{E}{-}3$} 
  & {\scriptsize$\pm4.87\mathrm{E}{-}4$} 
  & {\scriptsize$\pm4.17\mathrm{E}{-}3$} 
  & {\scriptsize$\mathbf{\pm1.58\mathrm{E}{-}3}$} 
  & {\scriptsize$\pm1.65\mathrm{E}{-}1$} 
  & {\scriptsize$\pm7.72\mathrm{E}{-}3$} \\
\bottomrule
\end{tabular}}
\label{tab:MSE}
\end{table}

For NSInv, Table~\ref{tab:nsinv_beta_errors}
reports the MSE of the inferred coefficients $\beta_1$ and $\beta_2$.
Relative to the baseline PINN, both P-PINN and D-PINN substantially improve
parameter recovery: P-PINN reduces the MSE of $\beta_1$ and $\beta_2$ by
approximately factors of four and thirteen, respectively, while D-PINN achieves the
smallest parameter errors overall. Taken together with the state errors in
Tables~\ref{tab:L2RE}–\ref{tab:MSE}, these results indicate that dropout
regularization can be particularly effective for this specific problem, whereas
P-PINN offers a more balanced trade-off between state and parameter accuracy
across different PDE settings.

\begin{table}[htbp]
\centering
\caption{Mean $\pm$ std of MSE of $\beta_1$ and $\beta_2$ for NSInv.}
\scalebox{0.8}{
\begin{tabular}{c|cc}
\toprule
\textbf{Method} & $\boldsymbol{\beta_1}$ \textbf{error} & $\boldsymbol{\beta_2}$ \textbf{error} \\
\midrule
PINN   & $1.79\mathrm{E}{-}1 \;\pm 1.04\mathrm{E}{-}2$ & $9.03\mathrm{E}{-}3 \;\pm 2.10\mathrm{E}{-}4$ \\
\midrule
P-PINN & $4.56\mathrm{E}{-}2 \;\pm 1.19\mathrm{E}{-}2$ & $7.05\mathrm{E}{-}4 \;\pm 4.88\mathrm{E}{-}4$ \\
\midrule
B-PINN   & $2.46\mathrm{E}{-}1 \;\pm 8.49\mathrm{E}{-}2$ & $5.02\mathrm{E}{-}4 \;\pm 4.54\mathrm{E}{-}4$ \\
\midrule
D-PINN & $\mathbf{2.23\mathrm{E}{-}2 \;\pm 4.84\mathrm{E}{-}3}$ & $\mathbf{1.03\mathrm{E}{-}4 \;\pm 2.97\mathrm{E}{-}6}$ \\
\bottomrule
\end{tabular}}
\label{tab:nsinv_beta_errors}
\end{table}

To further characterize approximation quality, Table~\ref{tab:MAX} reports
the mean and standard deviation of the MAE, while
Tables~\ref{tab:FMSE-L}–\ref{tab:FMSE-H} summarize the low-, mid-, and
high-frequency Fourier errors $\mathrm{fMSE}_\mathrm{L}$,
$\mathrm{fMSE}_\mathrm{M}$, and $\mathrm{fMSE}_\mathrm{H}$. P-PINN reduces the
maximum error on all inverse problems,
often by factors of $3$--$5$. In the spectral domain, P-PINN markedly lowers
$\mathrm{fMSE}_\mathrm{H}$ on the data
assimilation tasks and also decreases $\mathrm{fMSE}_\mathrm{L}$ and
$\mathrm{fMSE}_\mathrm{M}$. By contrast, B-PINN and D-PINN frequently amplify
low- or high-frequency components on several benchmarks (e.g., Poisson, PInv,
NSInv). These observations indicate that the
pruning-and-fine-tuning mechanism in P-PINN tends to selectively remove
noise-dominated, oscillatory components of the learned representation while
preserving and sharpening the physically meaningful low-frequency structure.

\begin{table}[H]
\centering
\caption{Mean $\pm$ std of MAE.}
\scalebox{0.8}{
\setlength{\tabcolsep}{2pt}
\begin{tabular}{c|cccc|ccccc}
\toprule
\textbf{Max error} & \multicolumn{4}{c|}{\textbf{Data assimilation}} & \multicolumn{5}{c}{\textbf{Parameter inversion}} \\
\midrule
                   & Poisson & Heat & Wave & Stokes & PInv & HInv & NSInv & EBInv & WInv \\
\midrule
\multirow{2}{*}{PINN}
  & $1.66\mathrm{E}{+}0$ & $1.08\mathrm{E}{+}0$ & $2.42\mathrm{E}{-}1$ & $3.32\mathrm{E}{+}0$ 
  & $7.89\mathrm{E}{-}2$ & $8.49\mathrm{E}{-}1$ & $3.46\mathrm{E}{+}0$ & $2.50\mathrm{E}{-}1$ & $1.99\mathrm{E}{-}1$ \\
  & {\scriptsize$\pm3.01\mathrm{E}{-}1$} 
  & {\scriptsize$\pm3.54\mathrm{E}{-}1$} 
  & {\scriptsize$\pm1.15\mathrm{E}{-}1$} 
  & {\scriptsize$\pm1.52\mathrm{E}{-}2$} 
  & {\scriptsize$\pm2.06\mathrm{E}{-}2$} 
  & {\scriptsize$\pm6.03\mathrm{E}{-}2$} 
  & {\scriptsize$\pm1.03\mathrm{E}{+}0$} 
  & {\scriptsize$\pm1.96\mathrm{E}{-}2$} 
  & {\scriptsize$\pm5.11\mathrm{E}{-}3$} \\
\midrule
\multirow{2}{*}{P-PINN}
  & $\mathbf{4.73\mathrm{E}{-}1}$ & $\mathbf{4.28\mathrm{E}{-}2}$ & $\mathbf{3.80\mathrm{E}{-}2}$ & $\mathbf{6.00\mathrm{E}{-}1}$ 
& $\mathbf{2.97\mathrm{E}{-}2}$ & $\mathbf{5.81\mathrm{E}{-}1}$ & $\mathbf{5.19\mathrm{E}{-}1}$ & $\mathbf{6.95\mathrm{E}{-}2}$ & $\mathbf{1.14\mathrm{E}{-}1}$ \\
& {\scriptsize$\mathbf{\pm1.32\mathrm{E}{-}1}$} 
& {\scriptsize$\mathbf{\pm3.04\mathrm{E}{-}2}$} 
& {\scriptsize$\mathbf{\pm4.22\mathrm{E}{-}3}$} 
& {\scriptsize$\mathbf{\pm1.28\mathrm{E}{-}1}$} 
& {\scriptsize$\mathbf{\pm5.86\mathrm{E}{-}3}$} 
& {\scriptsize$\mathbf{\pm2.54\mathrm{E}{-}2}$} 
& {\scriptsize$\mathbf{\pm5.69\mathrm{E}{-}2}$} 
& {\scriptsize$\mathbf{\pm7.14\mathrm{E}{-}3}$} 
& {\scriptsize$\mathbf{\pm6.91\mathrm{E}{-}3}$} \\ 
\midrule
\multirow{2}{*}{B-PINN}
  & $1.09\mathrm{E}{+}0$ & $4.34\mathrm{E}{-}1$ & $7.96\mathrm{E}{-}1$ & $2.11\mathrm{E}{+}0$ 
  & $5.63\mathrm{E}{-}1$ & $4.48\mathrm{E}{-}1$ & $3.86\mathrm{E}{+}0$ & $7.82\mathrm{E}{-}1$ & $4.25\mathrm{E}{-}1$ \\
  & {\scriptsize$\pm2.11\mathrm{E}{-}1$} 
  & {\scriptsize$\pm1.75\mathrm{E}{-}1$} 
  & {\scriptsize$\pm1.09\mathrm{E}{-}1$} 
  & {\scriptsize$\pm2.47\mathrm{E}{-}1$} 
  & {\scriptsize$\pm4.26\mathrm{E}{-}2$} 
  & {\scriptsize$\pm8.32\mathrm{E}{-}2$} 
  & {\scriptsize$\pm1.81\mathrm{E}{-}1$} 
  & {\scriptsize$\pm1.98\mathrm{E}{-}2$} 
  & {\scriptsize$\pm3.03\mathrm{E}{-}1$} \\
  \midrule
\multirow{2}{*}{D-PINN}
  & $1.15\mathrm{E}{+}0$ & $5.68\mathrm{E}{-}1$ & $2.09\mathrm{E}{-}1$ & $2.63\mathrm{E}{+}0$ 
  & $1.39\mathrm{E}{-}1$ & $3.63\mathrm{E}{-}1$ & $6.97\mathrm{E}{-}1$ & $3.90\mathrm{E}{-}1$ & $3.14\mathrm{E}{-}1$ \\
  & {\scriptsize$\pm2.54\mathrm{E}{-}2$} 
  & {\scriptsize$\pm3.50\mathrm{E}{-}2$} 
  & {\scriptsize$\pm1.94\mathrm{E}{-}2$} 
  & {\scriptsize$\pm9.11\mathrm{E}{-}2$} 
  & {\scriptsize$\pm1.38\mathrm{E}{-}2$} 
  & {\scriptsize$\pm4.11\mathrm{E}{-}3$} 
  & {\scriptsize$\pm4.81\mathrm{E}{-}2$} 
  & {\scriptsize$\pm2.83\mathrm{E}{-}1$} 
  & {\scriptsize$\pm2.79\mathrm{E}{-}2$} \\
\bottomrule
\end{tabular}}
\label{tab:MAX}
\end{table}

\begin{table}[htbp]
\centering
\caption{Mean $\pm$ std of fMSE-L.}
\scalebox{0.8}{
\setlength{\tabcolsep}{2pt}
\begin{tabular}{c|cccc|ccccc}
\toprule
\textbf{fMSE-L} & \multicolumn{4}{c|}{\textbf{Data assimilation}} & \multicolumn{5}{c}{\textbf{Parameter inversion}} \\
\midrule
                   & Poisson & Heat & Wave & Stokes & PInv & HInv & NSInv & EBInv & WInv  \\
\midrule
\multirow{2}{*}{PINN}
  & $2.80\mathrm{E}{+}2$ & $5.50\mathrm{E}{+}2$ & $1.08\mathrm{E}{+}2$ & $8.29\mathrm{E}{+}2$
  & $3.12\mathrm{E}{+}1$ & $1.39\mathrm{E}{+}2$ & $1.20\mathrm{E}{+}3$ & --- & --- \\
  & {\scriptsize$\pm2.21\mathrm{E}{+}1$} 
  & {\scriptsize$\pm2.16\mathrm{E}{+}2$} 
  & {\scriptsize$\pm5.03\mathrm{E}{+}1$} 
  & {\scriptsize$\pm1.32\mathrm{E}{+}1$} 
  & {\scriptsize$\pm9.64\mathrm{E}{+}0$} 
  & {\scriptsize$\pm2.54\mathrm{E}{+}1$} 
  & {\scriptsize$\pm7.25\mathrm{E}{+}2$} 
  &      &      \\
\midrule
\multirow{2}{*}{P-PINN}
  & $\mathbf{1.11\mathrm{E}{+}2}$ & $\mathbf{1.66\mathrm{E}{+}1}$ & $\mathbf{1.54\mathrm{E}{+}1}$ & $\mathbf{1.12\mathrm{E}{+}2}$
& $\mathbf{1.04\mathrm{E}{+}1}$ & $\mathbf{5.89\mathrm{E}{+}1}$ & $4.20\mathrm{E}{+}2$ & --- & --- \\
& {\scriptsize$\mathbf{\pm4.52\mathrm{E}{+}1}$} 
& {\scriptsize$\mathbf{\pm9.11\mathrm{E}{+}0}$} 
& {\scriptsize$\mathbf{\pm7.15\mathrm{E}{-}1}$} 
& {\scriptsize$\mathbf{\pm1.73\mathrm{E}{+}1}$} 
& {\scriptsize$\mathbf{\pm1.34\mathrm{E}{+}0}$} 
& {\scriptsize$\mathbf{\pm1.22\mathrm{E}{+}1}$} 
& {\scriptsize$\pm1.17\mathrm{E}{+}2$} 
&      & \\
\midrule
\multirow{2}{*}{B-PINN}
  & $1.97\mathrm{E}{+}3$ & $3.88\mathrm{E}{+}2$ & $6.96\mathrm{E}{+}2$ & $2.59\mathrm{E}{+}3$
  & $4.92\mathrm{E}{+}4$ & $2.11\mathrm{E}{+}2$ & $2.81\mathrm{E}{+}3$ & --- & --- \\
  & {\scriptsize$\pm6.18\mathrm{E}{+}2$} 
  & {\scriptsize$\pm3.56\mathrm{E}{+}2$} 
  & {\scriptsize$\pm3.40\mathrm{E}{+}2$} 
  & {\scriptsize$\pm4.90\mathrm{E}{+}2$} 
  & {\scriptsize$\pm1.58\mathrm{E}{+}4$} 
  & {\scriptsize$\pm2.34\mathrm{E}{+}1$} 
  & {\scriptsize$\pm5.59\mathrm{E}{+}2$} 
  &      &      \\
  \midrule
  \multirow{2}{*}{D-PINN}
  & $1.18\mathrm{E}{+}3$ & $1.87\mathrm{E}{+}2$ & $2.07\mathrm{E}{+}1$ & $1.93\mathrm{E}{+}3$
  & $1.59\mathrm{E}{+}3$ & $2.38\mathrm{E}{+}2$ & $\mathbf{2.62\mathrm{E}{+}2}$ & --- & --- \\
  & {\scriptsize$\pm1.15\mathrm{E}{+}1$} 
  & {\scriptsize$\pm1.98\mathrm{E}{+}1$} 
  & {\scriptsize$\pm4.64\mathrm{E}{+}0$} 
  & {\scriptsize$\pm2.93\mathrm{E}{+}1$} 
  & {\scriptsize$\pm4.07\mathrm{E}{+}2$} 
  & {\scriptsize$\pm3.28\mathrm{E}{+}0$} 
  & {\scriptsize$\mathbf{\pm6.23\mathrm{E}{+}1}$} 
  &      &      \\
\bottomrule
\end{tabular}}
\label{tab:FMSE-L}
\end{table}

\begin{table}[htbp]
\centering
\caption{Mean $\pm$ std of fMSE-M.}
\scalebox{0.8}{
\setlength{\tabcolsep}{2pt}
\begin{tabular}{c|cccc|ccccc}
\toprule
\textbf{fMSE-M} & \multicolumn{4}{c|}{\textbf{Data assimilation}} & \multicolumn{5}{c}{\textbf{Parameter inversion}} \\
\midrule
                   & Poisson & Heat & Wave & Stokes & PInv & HInv & NSInv & EBInv & WInv  \\
\midrule
\multirow{2}{*}{PINN}
  & $7.45\mathrm{E}{+}1$ & $5.12\mathrm{E}{+}1$ & $6.77\mathrm{E}{+}0$ & $9.26\mathrm{E}{+}1$
  & $5.60\mathrm{E}{-}1$ & $3.94\mathrm{E}{+}0$ & $4.87\mathrm{E}{+}1$ & --- & --- \\
  & {\scriptsize$\pm5.82\mathrm{E}{+}0$} 
  & {\scriptsize$\pm2.09\mathrm{E}{+}1$} 
  & {\scriptsize$\pm2.54\mathrm{E}{+}0$} 
  & {\scriptsize$\pm4.38\mathrm{E}{-}1$} 
  & {\scriptsize$\pm1.76\mathrm{E}{-}1$} 
  & {\scriptsize$\pm5.12\mathrm{E}{-}1$} 
  & {\scriptsize$\pm5.16\mathrm{E}{+}0$} 
  &      &      \\
\midrule
\multirow{2}{*}{P-PINN}
  & $\mathbf{1.16\mathrm{E}{+}1}$ & $\mathbf{7.40\mathrm{E}{-}1}$ & $\mathbf{1.91\mathrm{E}{+}0}$ & $\mathbf{1.19\mathrm{E}{+}1}$
& $\mathbf{1.90\mathrm{E}{-}1}$ & $\mathbf{2.18\mathrm{E}{+}0}$ & $\mathbf{5.59\mathrm{E}{+}0}$ & --- & --- \\
& {\scriptsize$\mathbf{\pm6.25\mathrm{E}{+}0}$} 
& {\scriptsize$\mathbf{\pm4.54\mathrm{E}{-}1}$} 
& {\scriptsize$\mathbf{\pm1.82\mathrm{E}{-}1}$} 
& {\scriptsize$\mathbf{\pm2.25\mathrm{E}{+}0}$} 
& {\scriptsize$\mathbf{\pm5.87\mathrm{E}{-}2}$} 
& {\scriptsize$\mathbf{\pm5.48\mathrm{E}{-}2}$} 
& {\scriptsize$\mathbf{\pm9.33\mathrm{E}{-}1}$} 
&      & \\
\midrule
\multirow{2}{*}{B-PINN}
  & $1.85\mathrm{E}{+}2$ & $8.53\mathrm{E}{+}0$ & $5.42\mathrm{E}{+}0$ & $1.74\mathrm{E}{+}2$
  & $2.52\mathrm{E}{+}2$ & $1.55\mathrm{E}{+}1$ & $1.21\mathrm{E}{+}2$ & --- & --- \\
  & {\scriptsize$\pm6.24\mathrm{E}{+}1$} 
  & {\scriptsize$\pm6.50\mathrm{E}{+}0$} 
  & {\scriptsize$\pm2.00\mathrm{E}{+}0$} 
  & {\scriptsize$\pm1.62\mathrm{E}{+}1$} 
  & {\scriptsize$\pm4.98\mathrm{E}{+}1$} 
  & {\scriptsize$\pm3.50\mathrm{E}{+}0$} 
  & {\scriptsize$\pm1.08\mathrm{E}{+}1$} 
  &      &      \\
  \midrule
\multirow{2}{*}{D-PINN}
  & $5.35\mathrm{E}{+}1$ & $1.54\mathrm{E}{+}0$ & $2.12\mathrm{E}{-}1$ & $2.16\mathrm{E}{+}2$
  & $5.05\mathrm{E}{-}1$ & $2.73\mathrm{E}{+}0$ & $9.12\mathrm{E}{+}0$ & --- & --- \\
  & {\scriptsize$\pm1.22\mathrm{E}{+}1$} 
  & {\scriptsize$\pm3.84\mathrm{E}{-}1$} 
  & {\scriptsize$\pm3.52\mathrm{E}{-}2$} 
  & {\scriptsize$\pm3.85\mathrm{E}{+}0$} 
  & {\scriptsize$\pm7.52\mathrm{E}{-}2$} 
  & {\scriptsize$\pm5.70\mathrm{E}{-}3$} 
  & {\scriptsize$\pm1.51\mathrm{E}{-}1$} 
  &      &      \\
\bottomrule
\end{tabular}}
\label{tab:FMSE-M}
\end{table}
\begin{table}[htbp]
\centering
\caption{Mean $\pm$ std of fMSE-H.}
\scalebox{0.8}{
\setlength{\tabcolsep}{2pt}
\begin{tabular}{c|cccc|ccccc}
\toprule
\textbf{fMSE-H} & \multicolumn{4}{c|}{\textbf{Data assimilation}} & \multicolumn{5}{c}{\textbf{Parameter inversion}} \\
\midrule
                   & Poisson & Heat & Wave & Stokes & PInv & HInv & NSInv & EBInv & WInv  \\
\midrule
\multirow{2}{*}{PINN}
  & $7.47\mathrm{E}{+}0$ & $6.06\mathrm{E}{+}0$ & $7.15\mathrm{E}{-}1$ & $1.33\mathrm{E}{+}1$
  & $8.35\mathrm{E}{-}2$ & $5.41\mathrm{E}{-}1$ & $2.13\mathrm{E}{+}0$ & --- & --- \\
  & {\scriptsize$\pm3.85\mathrm{E}{-}1$} 
  & {\scriptsize$\pm2.92\mathrm{E}{+}0$} 
  & {\scriptsize$\pm3.23\mathrm{E}{-}1$} 
  & {\scriptsize$\pm9.74\mathrm{E}{-}2$} 
  & {\scriptsize$\pm2.25\mathrm{E}{-}2$} 
  & {\scriptsize$\pm7.56\mathrm{E}{-}2$} 
  & {\scriptsize$\pm1.94\mathrm{E}{-}1$} 
  &      &      \\
\midrule
\multirow{2}{*}{P-PINN}
  & $\mathbf{1.57\mathrm{E}{+}0}$ & $\mathbf{9.75\mathrm{E}{-}2}$ & $\mathbf{1.76\mathrm{E}{-}1}$ & $\mathbf{1.72\mathrm{E}{+}0}$
& $\mathbf{1.28\mathrm{E}{-}2}$ & $\mathbf{3.46\mathrm{E}{-}1}$ & $\mathbf{6.51\mathrm{E}{-}1}$ & --- & --- \\
& {\scriptsize$\mathbf{\pm8.52\mathrm{E}{-}1}$} 
& {\scriptsize$\mathbf{\pm5.64\mathrm{E}{-}2}$} 
& {\scriptsize$\mathbf{\pm2.05\mathrm{E}{-}2}$} 
& {\scriptsize$\mathbf{\pm3.06\mathrm{E}{-}1}$} 
& {\scriptsize$\mathbf{\pm9.72\mathrm{E}{-}4}$} 
& {\scriptsize$\mathbf{\pm5.04\mathrm{E}{-}2}$} 
& {\scriptsize$\mathbf{\pm1.47\mathrm{E}{-}1}$} 
&      &  \\
  \midrule
\multirow{2}{*}{B-PINN}
  & $1.35\mathrm{E}{+}1$ & $4.26\mathrm{E}{-}1$ & $2.70\mathrm{E}{-}1$ & $1.42\mathrm{E}{+}1$
  & $7.84\mathrm{E}{+}0$ & $3.66\mathrm{E}{+}0$ & $1.07\mathrm{E}{+}0$ & --- & --- \\
  & {\scriptsize$\pm4.54\mathrm{E}{+}0$} 
  & {\scriptsize$\pm3.23\mathrm{E}{-}1$} 
  & {\scriptsize$\pm1.01\mathrm{E}{-}1$} 
  & {\scriptsize$\pm1.15\mathrm{E}{+}0$} 
  & {\scriptsize$\pm1.73\mathrm{E}{+}0$} 
  & {\scriptsize$\pm7.83\mathrm{E}{-}1$} 
  & {\scriptsize$\pm9.24\mathrm{E}{-}1$} 
  &      &      \\
  \midrule
  \multirow{2}{*}{D-PINN}
  & $1.00\mathrm{E}{+}1$ & $5.05\mathrm{E}{-}1$ & $1.70\mathrm{E}{-}2$ & $1.69\mathrm{E}{+}1$
  & $2.16\mathrm{E}{-}1$ & $1.09\mathrm{E}{+}0$ & $1.28\mathrm{E}{+}0$ & --- & --- \\
  & {\scriptsize$\pm2.91\mathrm{E}{-}1$} 
  & {\scriptsize$\pm2.25\mathrm{E}{-}2$} 
  & {\scriptsize$\pm6.49\mathrm{E}{-}4$} 
  & {\scriptsize$\pm3.55\mathrm{E}{-}1$} 
  & {\scriptsize$\pm3.38\mathrm{E}{-}3$} 
  & {\scriptsize$\pm2.12\mathrm{E}{-}2$} 
  & {\scriptsize$\pm1.22\mathrm{E}{-}2$} 
  &      &      \\
\bottomrule
\end{tabular}}
\label{tab:FMSE-H}
\end{table}
\subsection{Qualitative reconstructions}
\label{subsec:qualitative}

\begin{figure}[!htbp]
\centering
\subfigure[PINN]{\includegraphics[width=0.32\textwidth]{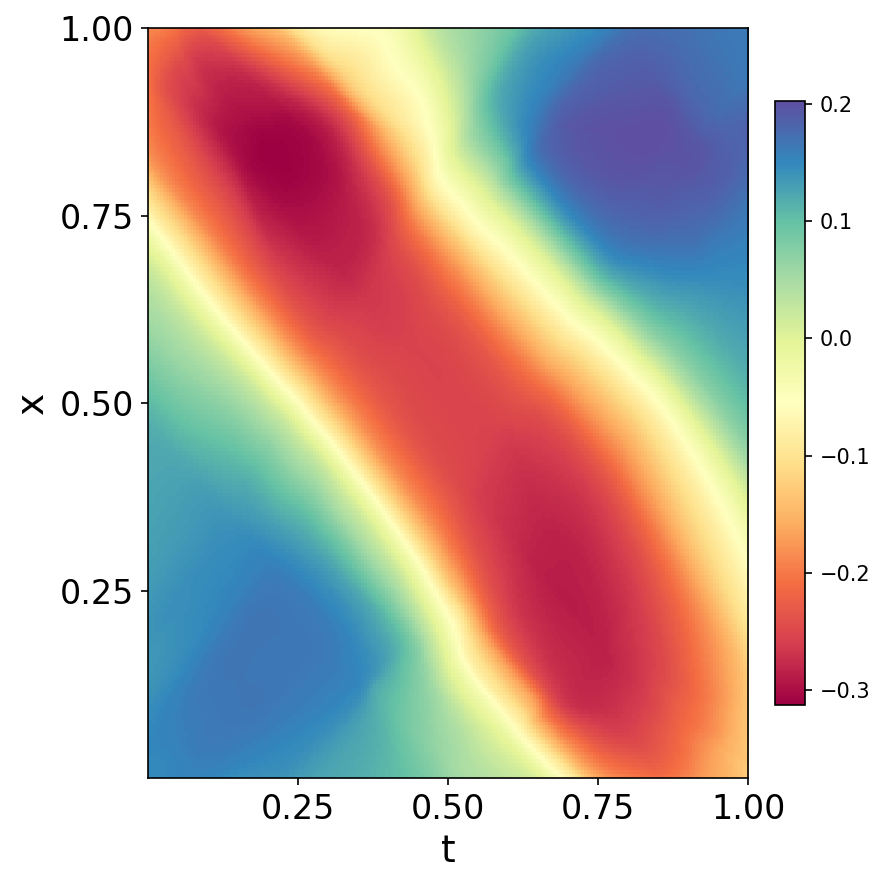}}
\subfigure[Exact]{\includegraphics[width=0.32\textwidth]{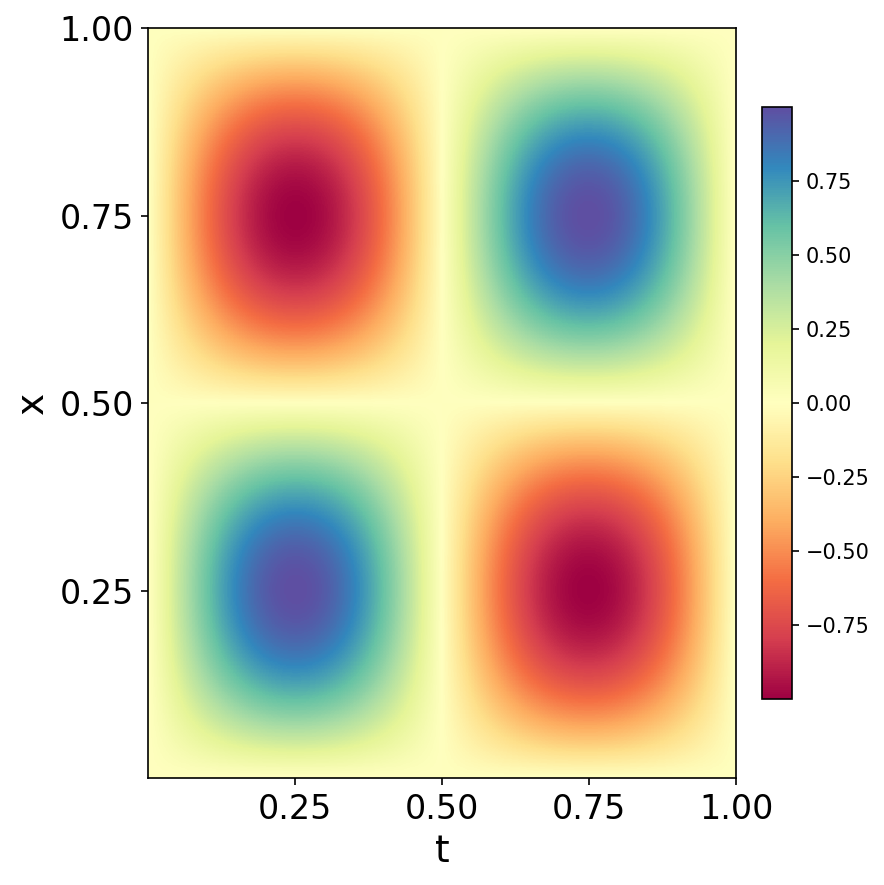}}
\subfigure[P-PINN]{\includegraphics[width=0.32\textwidth]{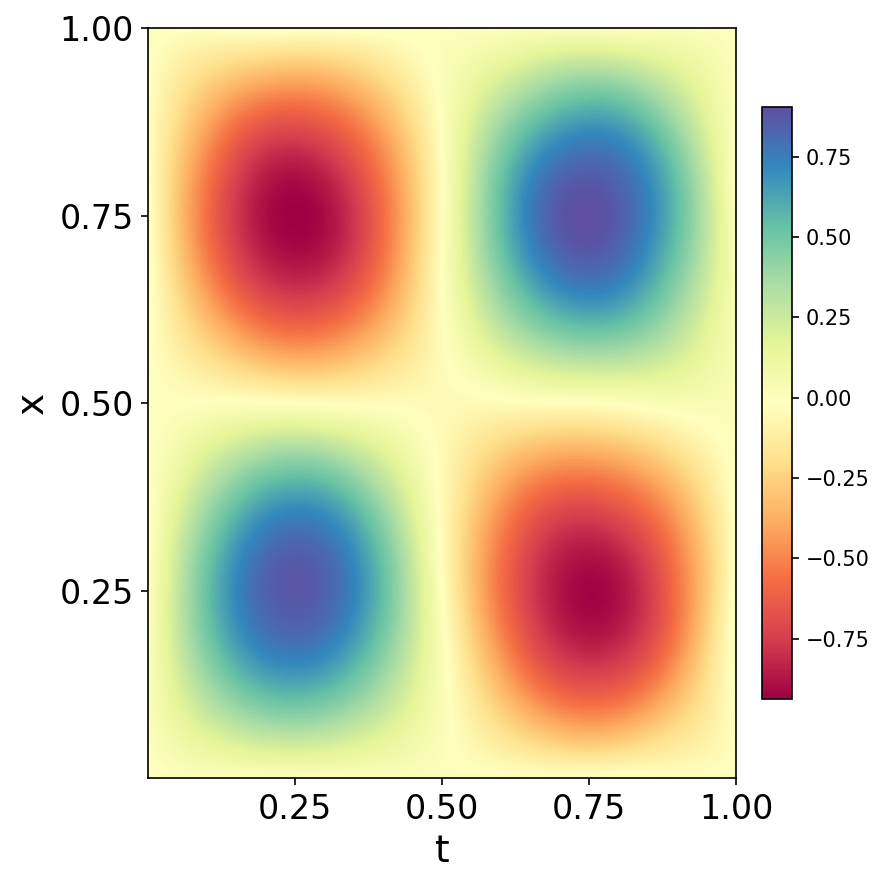}}
\caption{Solution fields for the one-dimensional wave data assimilation problem:
 (a) standard PINN prediction, which roughly identifies the correct locations of the wave lobes
  but is contaminated by pronounced spurious oscillations; (b) exact solution $u(x,t)$; (c) P-PINN prediction after selective pruning and fine-tuning, which closely matches the
  smooth standing-wave structure of the exact solution.}
\label{fig:data-w}
\end{figure}

\begin{figure}[!htbp]
    \centering
    \subfigure[PINN]{\includegraphics[width=0.32\textwidth]{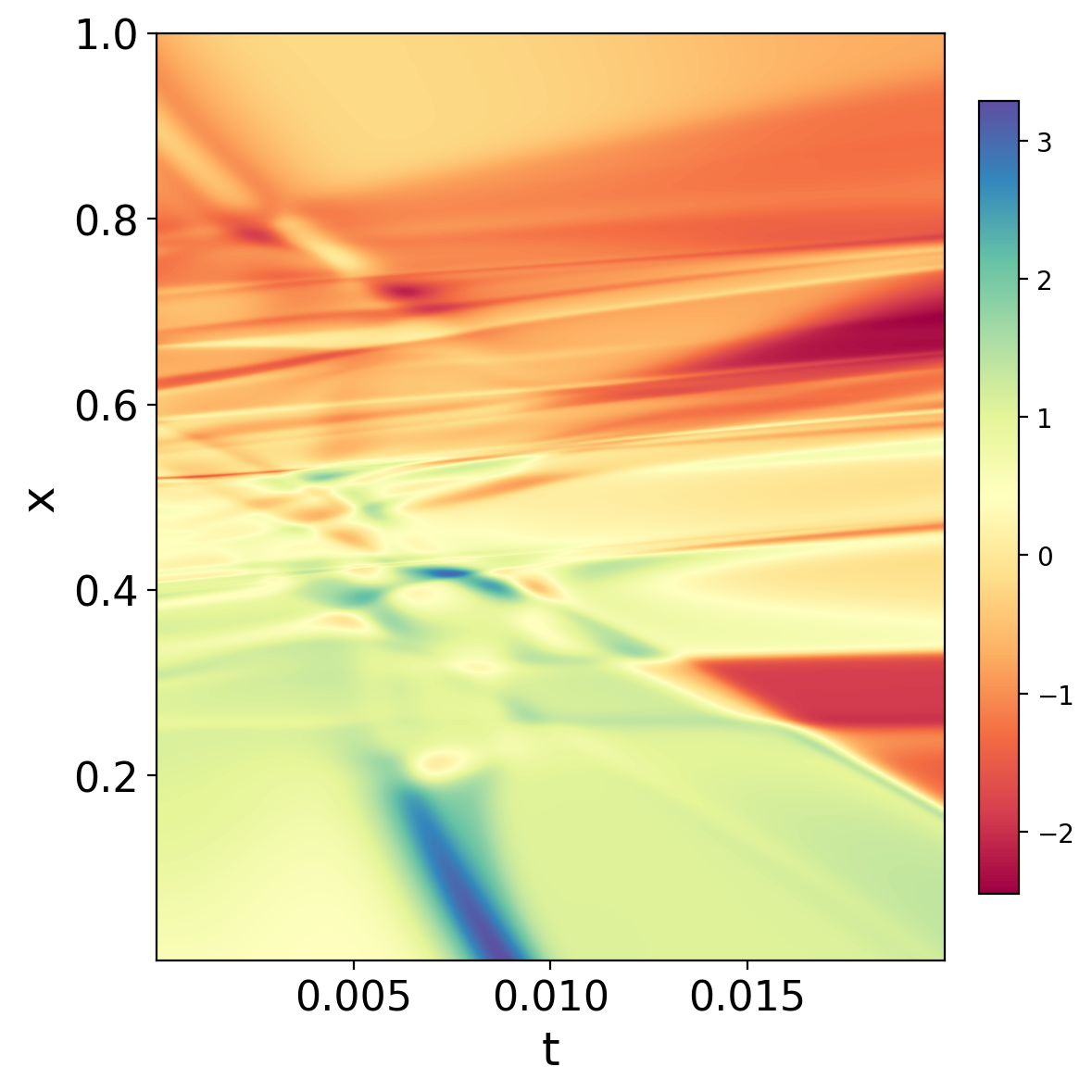}}
    \subfigure[Exact]{\includegraphics[width=0.32\textwidth]{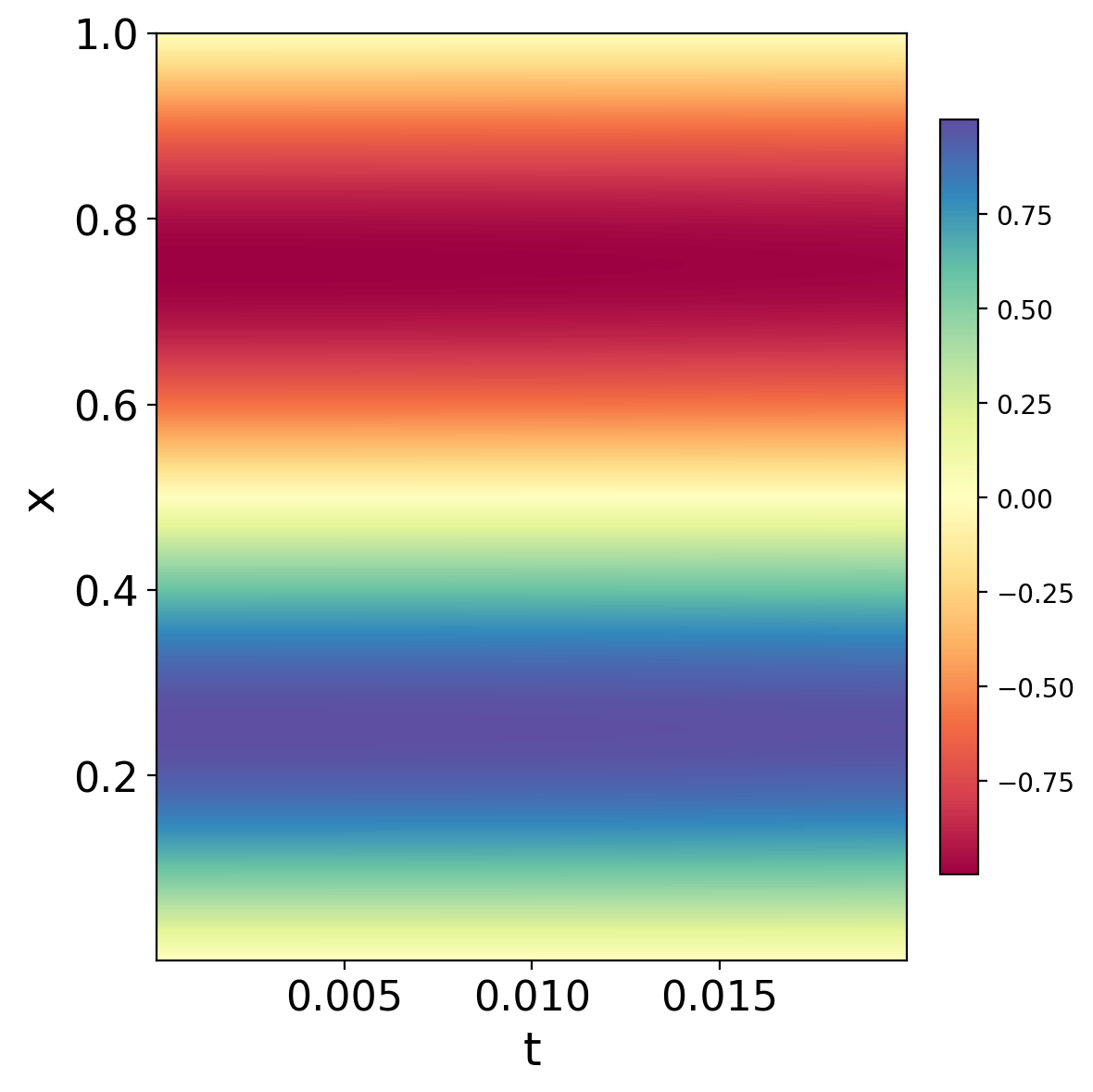}}
    \subfigure[P-PINN]{\includegraphics[width=0.32\textwidth]{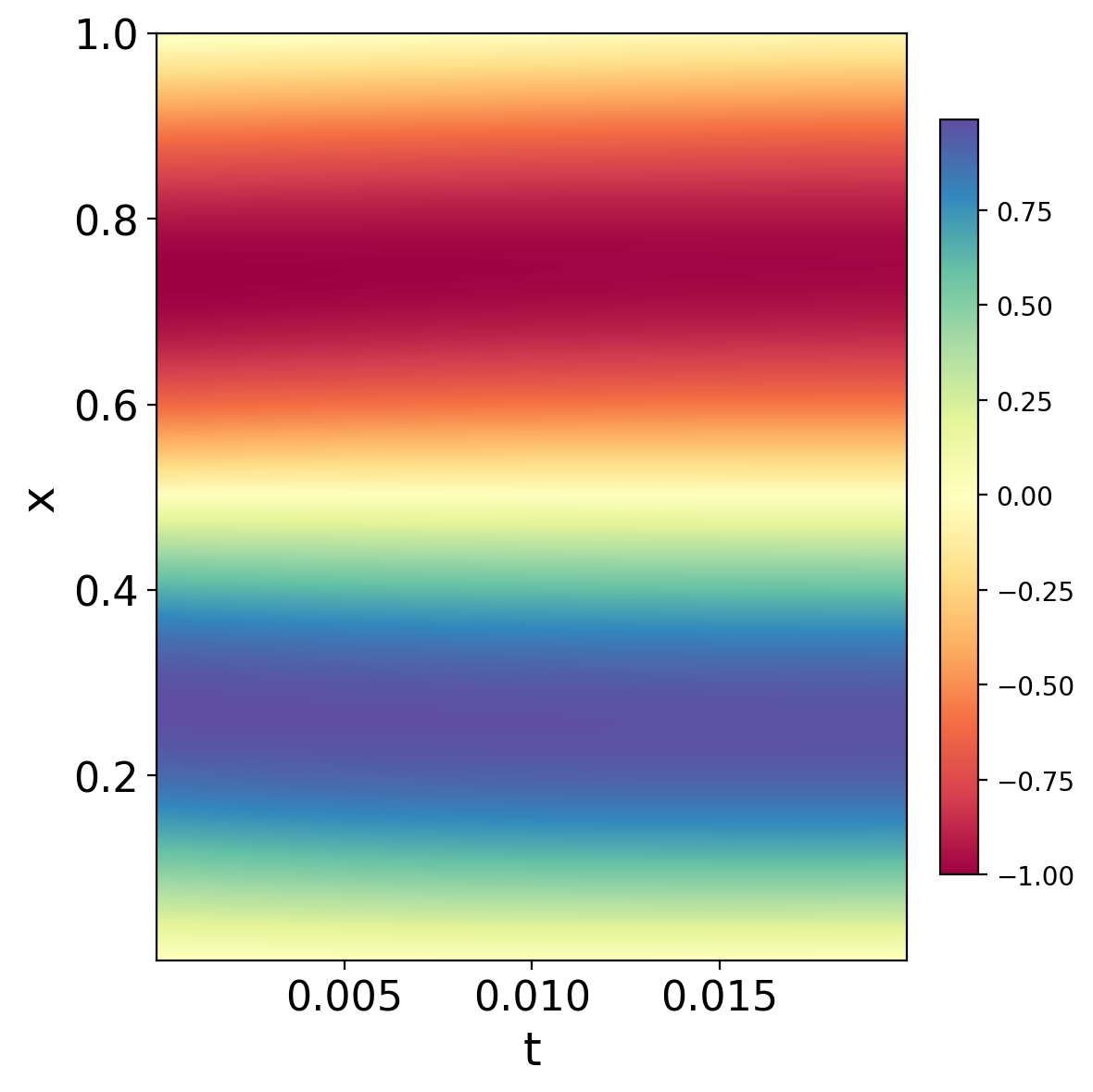}}
    \caption{Heat equation data assimilation: (a) standard PINN prediction;
    (b) exact solution $u(x,t)$; (c) P-PINN prediction after selective pruning
    and fine-tuning.}
    \label{fig:data-h}
\end{figure}

\begin{figure}[!htbp]
    \centering
    \subfigure[PINN]{\includegraphics[width=0.32\textwidth]{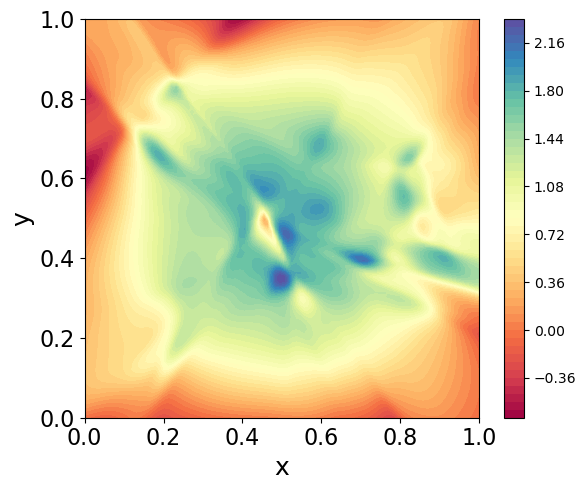}}
    \subfigure[Exact]{\includegraphics[width=0.32\textwidth]{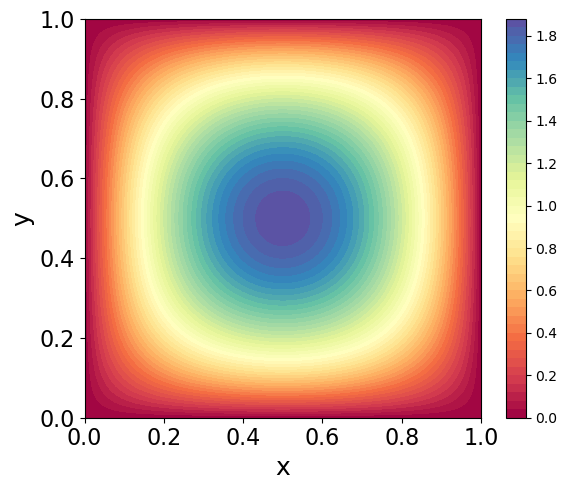}}
    \subfigure[P-PINN]{\includegraphics[width=0.32\textwidth]{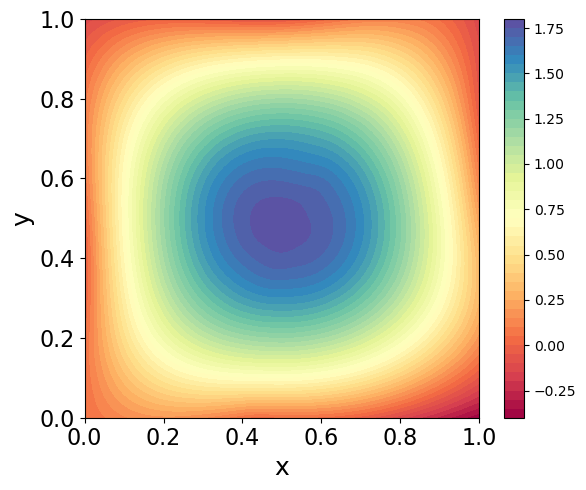}}
    \caption{Poisson data assimilation: (a) standard PINN prediction;
    (b) exact solution $u(x,y)$; (c) P-PINN prediction.}
    \label{fig:data-p}
\end{figure}

\begin{figure}[!htbp]
  \centering
  \subfigure[$u$-component, PINN]{\includegraphics[width=0.32\textwidth]{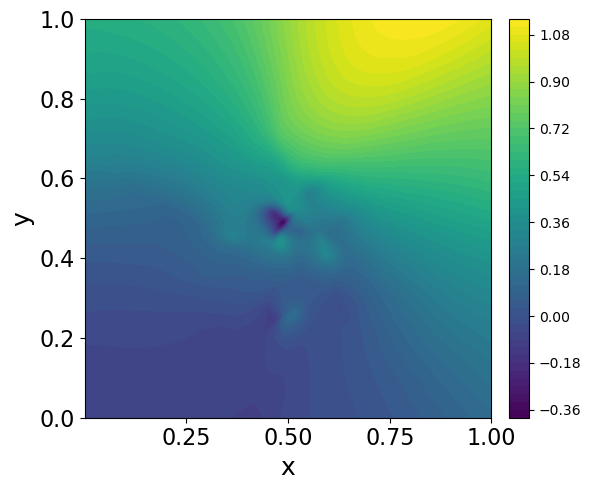}}
  \subfigure[$u$-component, exact]{\includegraphics[width=0.32\textwidth]{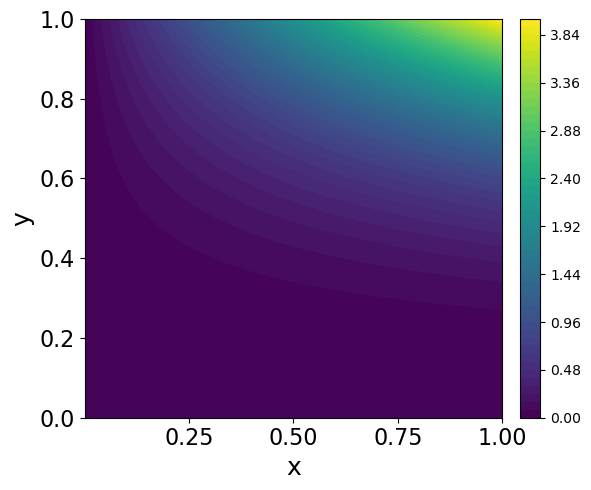}}
  \subfigure[$u$-component, P-PINN]{\includegraphics[width=0.32\textwidth]{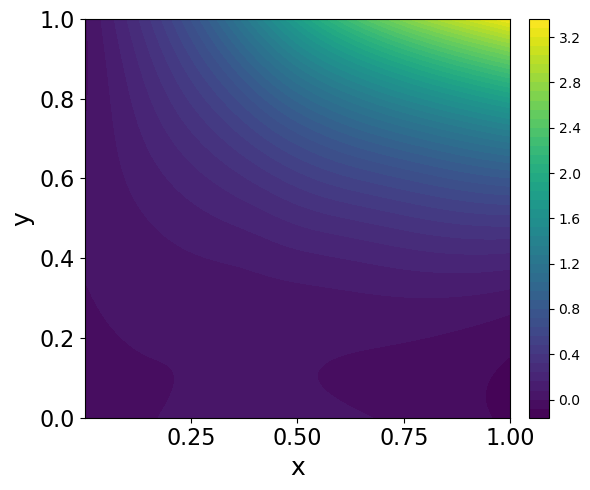}}\\
  \subfigure[$v$-component, PINN]{\includegraphics[width=0.32\textwidth]{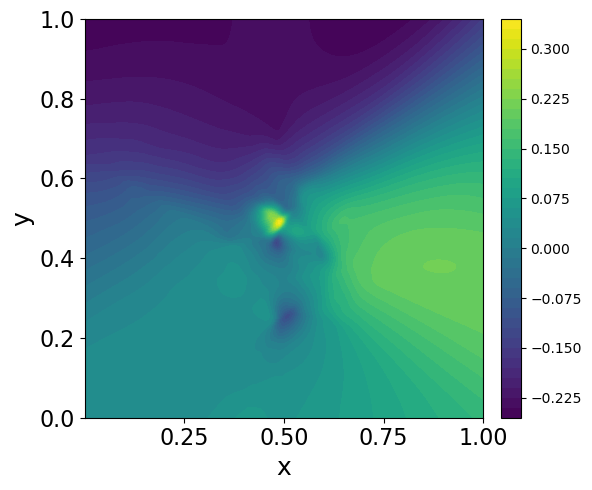}}
  \subfigure[$v$-component, exact]{\includegraphics[width=0.32\textwidth]{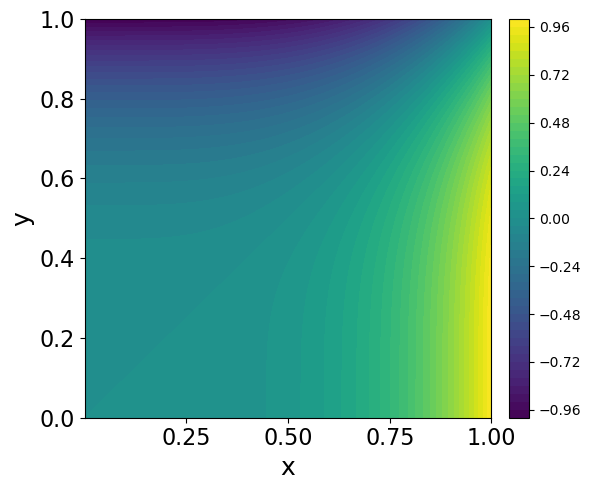}}
  \subfigure[$v$-component, P-PINN]{\includegraphics[width=0.32\textwidth]{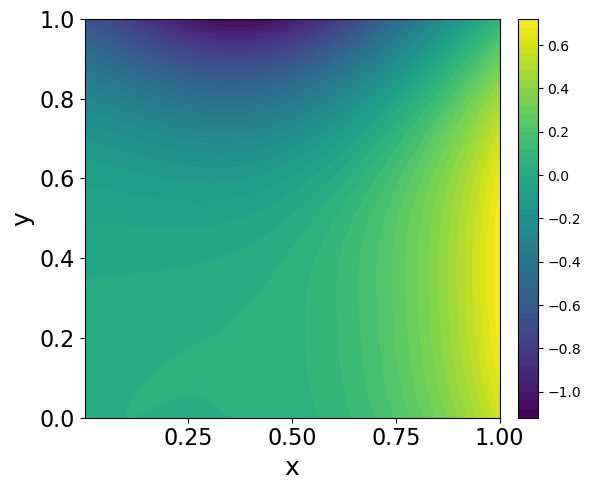}}\\
  \subfigure[$p$-component, PINN]{\includegraphics[width=0.32\textwidth]{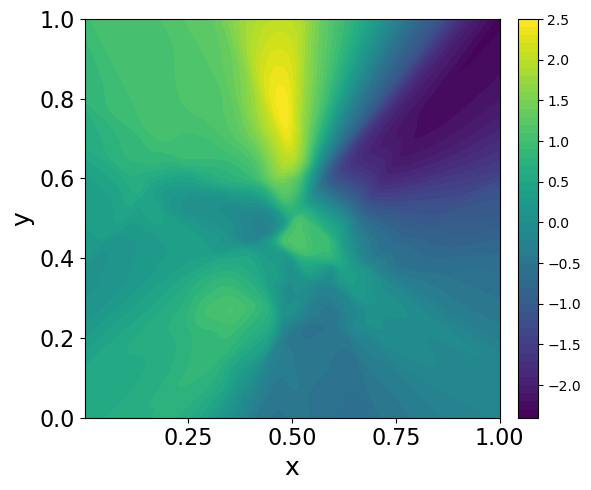}}
  \subfigure[$p$-component, exact]{\includegraphics[width=0.32\textwidth]{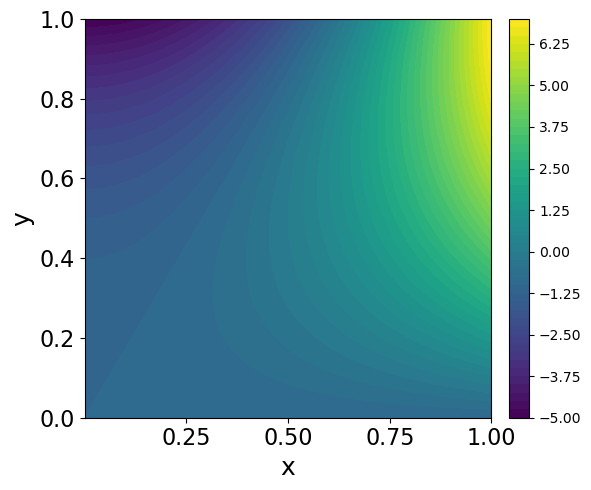}}
  \subfigure[$p$-component, P-PINN]{\includegraphics[width=0.32\textwidth]{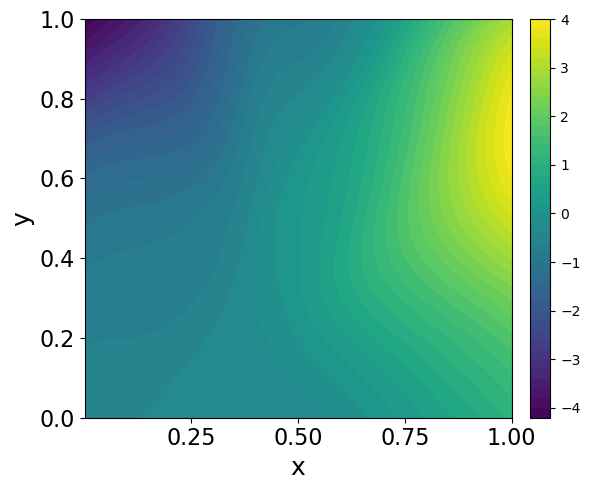}}
  \caption{Stokes data assimilation: velocity and pressure components.
    Rows from top to bottom: $u$-component, $v$-component, and pressure $p$.
    Columns from left to right: standard PINN prediction, exact solution,
    and P-PINN prediction.}
  \label{fig:data-s}
\end{figure}

\begin{figure}[!htbp]
  \centering
  \subfigure[PINN]{\includegraphics[width=0.32\textwidth]{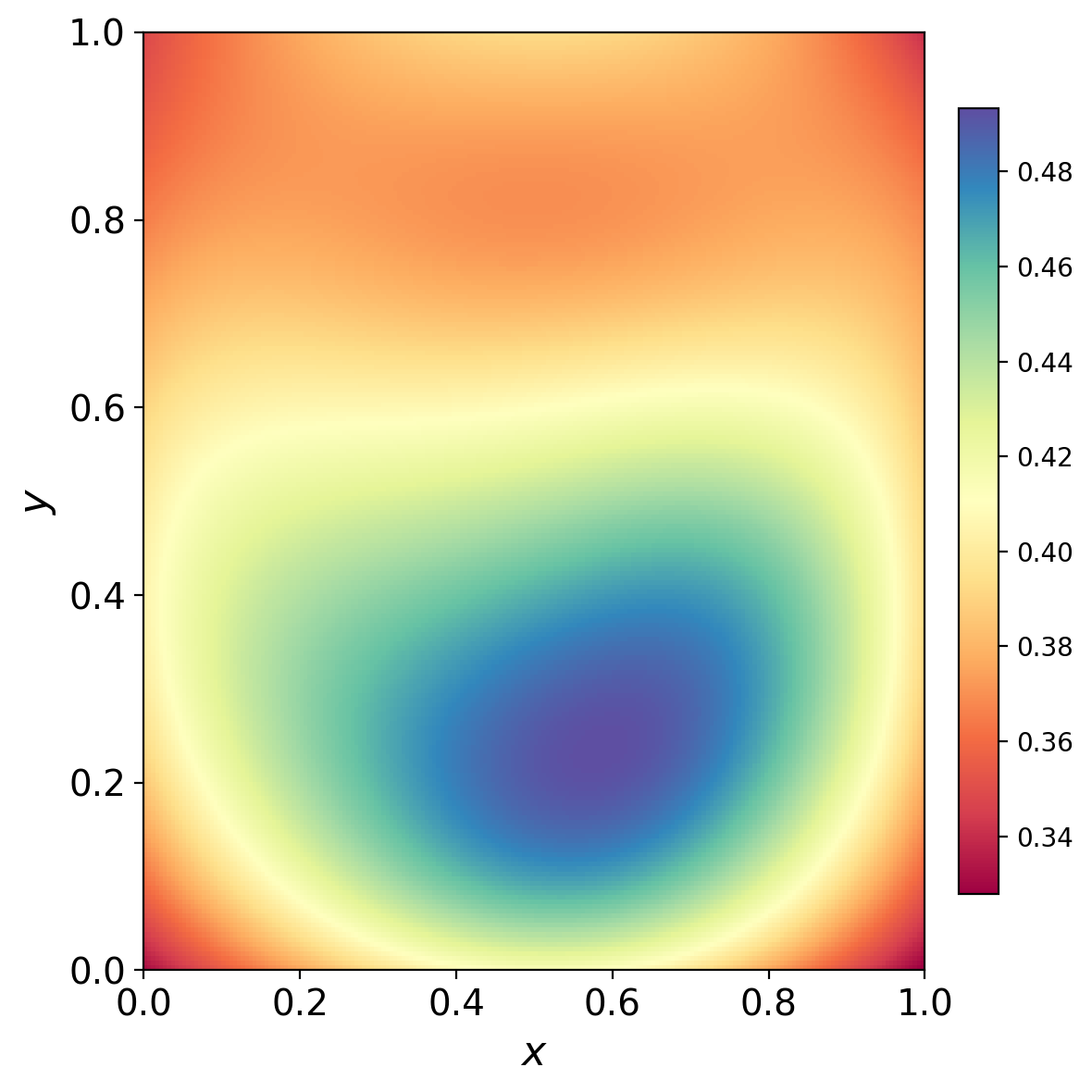}}
  \subfigure[Exact]{\includegraphics[width=0.32\textwidth]{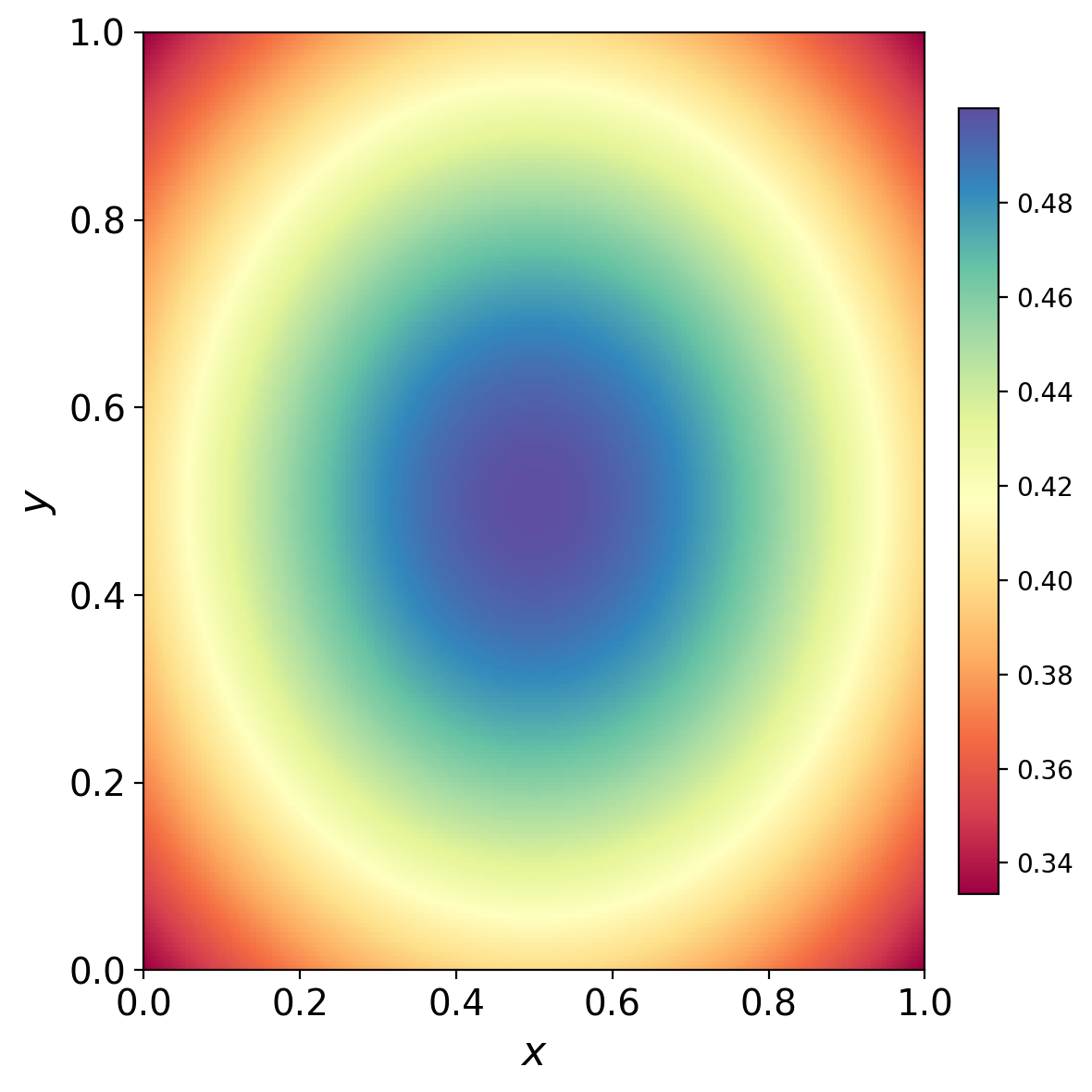}}
  \subfigure[P-PINN]{\includegraphics[width=0.32\textwidth]{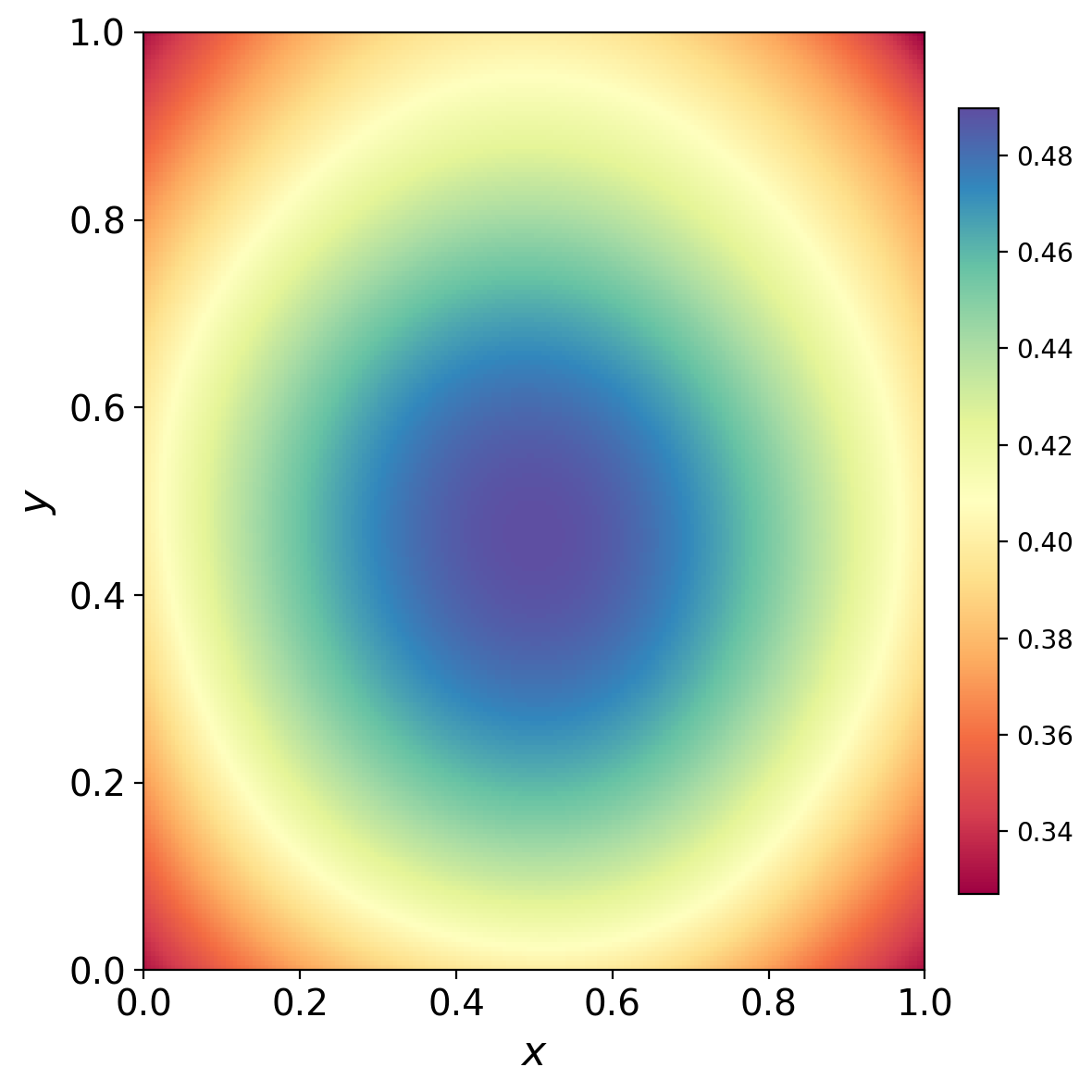}}
  \caption{Diffusion coefficient $a(x,y)$ for the Poisson inverse problem:
  (a) standard PINN prediction; (b) ground truth; (c) P-PINN prediction.}
  \label{fig:PINV}
\end{figure}

\begin{figure}[!htbp]
  \centering
  \subfigure[PINN]{\includegraphics[width=0.32\textwidth]{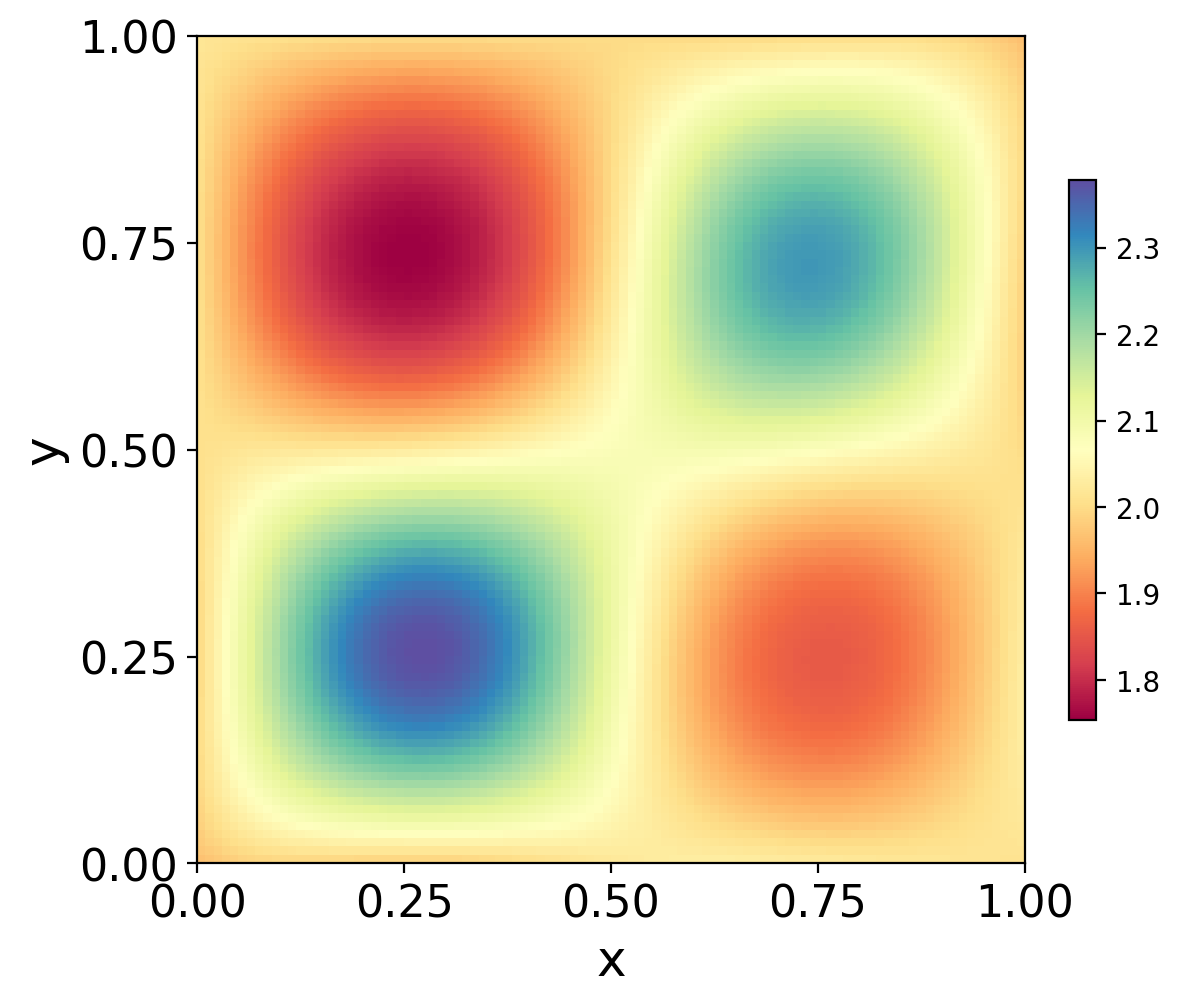}}
  \subfigure[Exact]{\includegraphics[width=0.32\textwidth]{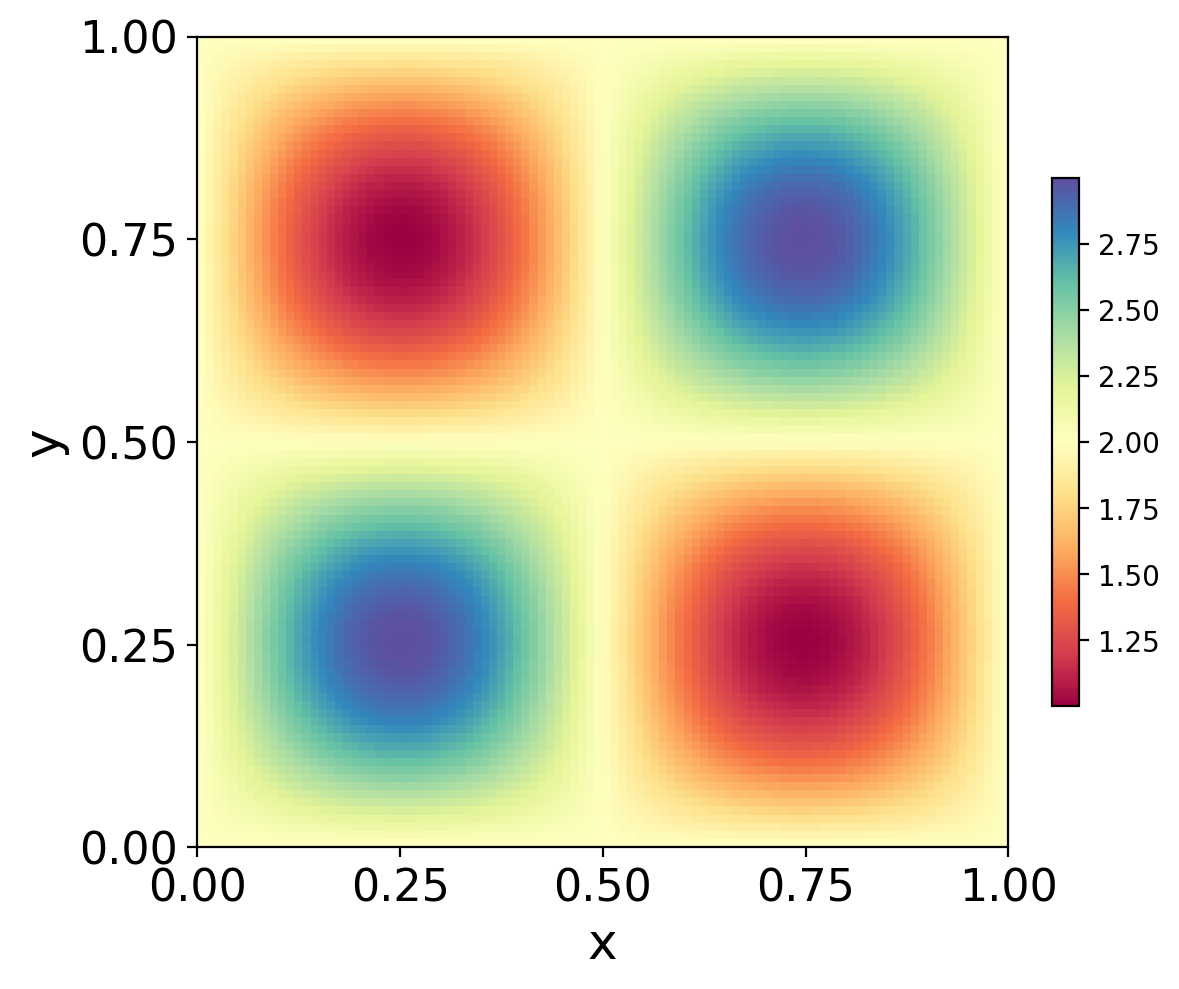}}
  \subfigure[P-PINN]{\includegraphics[width=0.32\textwidth]{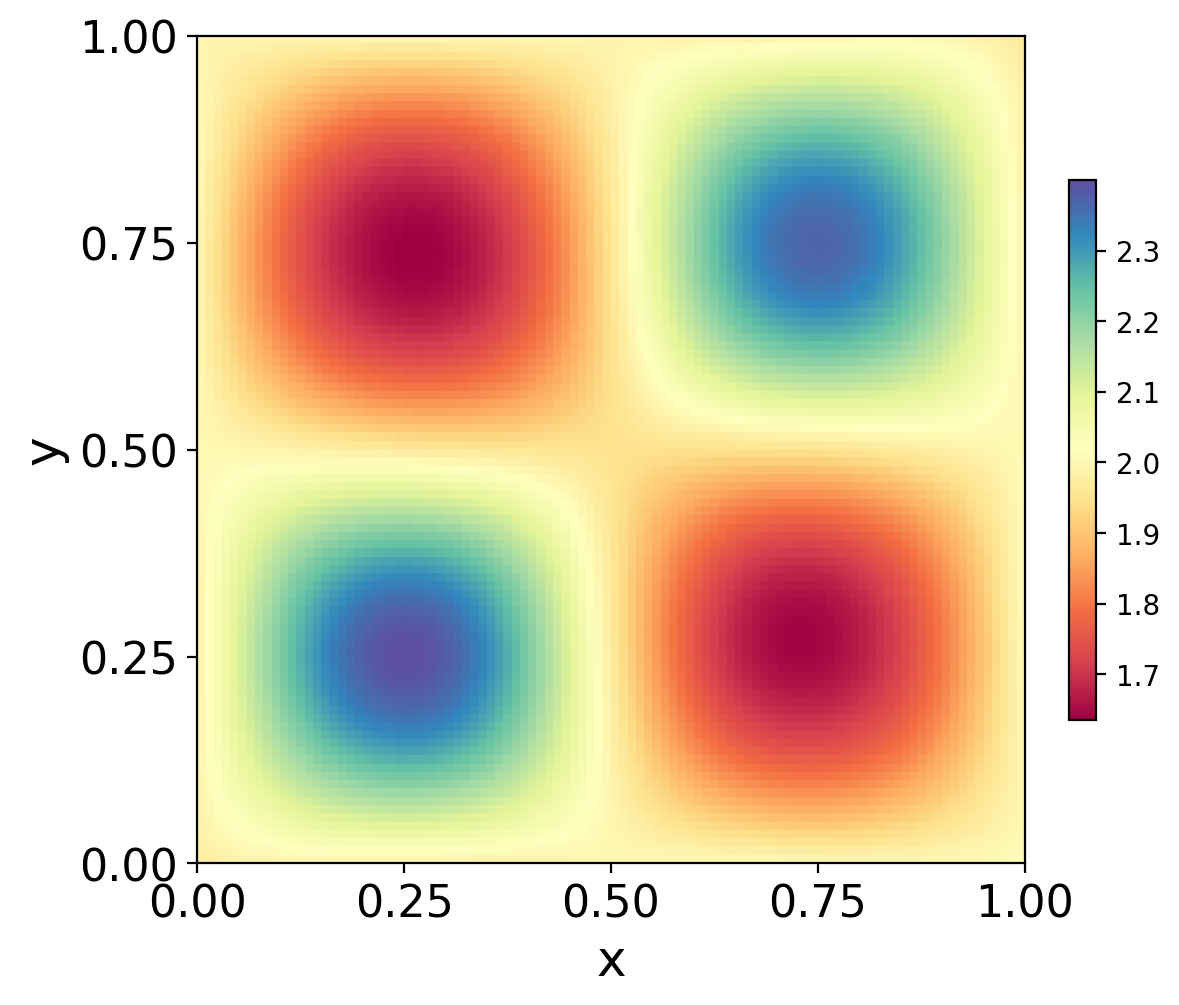}}
  \caption{Diffusion coefficient $a(x,y)$ for the heat inverse problem:
  (a) standard PINN prediction; (b) ground truth; (c) P-PINN prediction.}
  \label{fig:HINV}
\end{figure}

We now illustrate the qualitative behavior of P-PINN on representative
inverse and data assimilation problems. Figures~\ref{fig:data-w}--\ref{fig:HINV}
compare baseline PINN reconstructions, exact fields, and P-PINN reconstructions.

Figure~\ref{fig:data-w} shows solution fields for the one-dimensional wave data
assimilation problem. The baseline PINN (panel (a)) exhibits spurious
oscillations and phase errors, particularly away from the observation regions,
whereas P-PINN (panel (c)) closely follows the exact solution (panel (b)) and
substantially reduces these artifacts.

Figures~\ref{fig:data-h} and~\ref{fig:data-p} present analogous comparisons for
the heat and Poisson data assimilation problems, respectively. In both cases,
the baseline PINN reconstructions are visibly contaminated by the noisy observations,
while P-PINN produces smoother fields that agree more closely with the exact solutions
and respect the underlying PDE structure.

Figure~\ref{fig:data-s} shows velocity and pressure reconstructions for the
Stokes data assimilation problem. The baseline PINN displays noticeable
violations of incompressibility and noisy pressure oscillations, whereas
P-PINN achieves smoother velocity fields and a more regular pressure
distribution that is visually closer to the reference fields.

Figures~\ref{fig:PINV}--\ref{fig:HINV} show representative parameter-inversion reconstructions. For both PInv and HInv, P-PINN recovers coefficient fields that are visually closer to the ground truth and exhibit markedly fewer noise-driven artifacts than the baseline PINN. For NSInv, P-PINN also improves the recovered pressure field; see the right panel of Fig.~\ref{fig:ppinn-framework-ns} for a direct comparison.

\vspace*{-0.4\baselineskip}
\subsection{Computational efficiency and pruning design}
\label{subsec:training-efficiency}

We next compare the computational cost and accuracy of the three pipelines that
operate after data partitio\-ning---retraining from scratch on $\mathcal{D}_{\text{retain}}$ (RT), naive fine-tuning
of the baseline PINN on $\mathcal{D}_{\text{retain}}$ without pruning (FT), and
the proposed P-PINN, which combines selective pruning with fine-tuning. All
methods start from the same partition
$\mathcal{D}_{\text{retain}}/\mathcal{D}_{\text{forget}}$ constructed by the
composite score in Section~\ref{subsec:partition}. Table~\ref{tab:time_l2re}
reports the total wall-clock training time and the resulting test L2RE on four
representative PDE problems (Heat, Wave, PInv, NSInv), measured on a single
NVIDIA A100 GPU.

\begin{table}[htbp]
\caption{Training time and test L2RE for three strategies across four PDE problems.}
\centering
\label{tab:time_l2re}
\begingroup
\setlength{\tabcolsep}{2pt}
\small

\begin{tabular}{@{}c|ccc@{}}
\multicolumn{4}{c}{(a) Time (s)}\\[2pt]
\toprule
\textbf{Problem} & \textbf{RT} & \textbf{FT} & \textbf{P-PINN} \\
\midrule
Heat   & $7.24\mathrm{E}{+}1$ & $1.12\mathrm{E}{+}1$ & $1.62\mathrm{E}{+}1$ \\
Wave   & $9.98\mathrm{E}{+}1$ & $1.97\mathrm{E}{+}1$ & $2.00\mathrm{E}{+}1$ \\
PInv   & $1.65\mathrm{E}{+}2$ & $1.69\mathrm{E}{+}1$ & $1.70\mathrm{E}{+}1$ \\
NSInv  & $3.22\mathrm{E}{+}2$ & $3.03\mathrm{E}{+}1$ & $3.18\mathrm{E}{+}1$ \\
\bottomrule
\end{tabular}
\hspace{0.05\textwidth}
\begin{tabular}{@{}c|ccc@{}}
\multicolumn{4}{c}{(b) L2RE}\\[2pt]
\toprule
\textbf{Problem} & \textbf{RT} & \textbf{FT} & \textbf{P-PINN} \\
\midrule
Heat   & $7.66\mathrm{E}{-}1$ & $1.80\mathrm{E}{+}0$ & $3.29\mathrm{E}{-}2$ \\
Wave   & $2.43\mathrm{E}{-}1$ & $8.05\mathrm{E}{-}1$ & $1.07\mathrm{E}{-}1$ \\
PInv   & $1.92\mathrm{E}{-}2$ & $2.39\mathrm{E}{-}2$ & $1.61\mathrm{E}{-}2$ \\
NSInv  & $2.30\mathrm{E}{+}0$ & $1.61\mathrm{E}{+}0$ & $4.91\mathrm{E}{-}1$ \\
\bottomrule
\end{tabular}

\endgroup
\end{table}
\vspace*{-0.8\baselineskip}
Table~\ref{tab:time_l2re} indicates that P-PINN achieves lower L2RE than both retraining and naive fine-tuning on all
four problems, while incurring only a modest overhead relative to FT and
requiring substantially less time than RT. In particular, on Heat and PInv,
P-PINN attains the best accuracy with a training cost comparable to fine-tuning
and more than an order of magnitude cheaper than full retraining. These results
demonstrate that the pruning-and-fine-tuning pipeline can improve accuracy
without substantially increasing computational cost.

We then analyze how the pruning strategy and neuron-importance criterion affect
performance. Table~\ref{tab:strategy_l2re} compares single-step and iterative
pruning under the same overall sparsity and retention ratio for four PDE
benchmarks (Heat, Wave, PInv, NSInv). Iterative pruning consistently
outperforms single-step pruning, especially on harder problems such as NSInv
and Heat, indicating the benefit of recomputing importance scores after each
pruning stage.

\begin{table}[htbp]
\centering
\caption{Effect of pruning strategies on L2RE (mean $\pm$ std) across four PDEs.}
\label{tab:strategy_l2re}

\setlength{\tabcolsep}{3pt}
\renewcommand{\arraystretch}{1.0}
\scalebox{0.8}{
\begin{tabular}{c|cccc}
\toprule
\textbf{Strategy} & \textbf{Heat} & \textbf{Wave} & \textbf{PInv} & \textbf{NSInv} \\
\midrule
Iterative &
$\mathbf{3.26\mathrm{E}{-}2}$ & $\mathbf{8.88\mathrm{E}{-}2}$ & $\mathbf{1.82\mathrm{E}{-}2}$ & $\mathbf{6.44\mathrm{E}{-}1}$ \\
& {\scriptsize$\pm1.56\mathrm{E}{-}4$} & {\scriptsize$\pm4.38\mathrm{E}{-}4$} &
  {\scriptsize$\pm2.17\mathrm{E}{-}3$} & {\scriptsize$\pm2.67\mathrm{E}{-}1$} \\
\midrule
Single &
$7.28\mathrm{E}{-}1$ & $\mathbf{8.81\mathrm{E}{-}2}$ & $1.94\mathrm{E}{-}2$ & $8.87\mathrm{E}{-}1$ \\
& {\scriptsize$\pm1.77\mathrm{E}{-}2$} & {\scriptsize$\pm6.05\mathrm{E}{-}4$} &
  {\scriptsize$\pm4.67\mathrm{E}{-}3$} & {\scriptsize$\pm1.32\mathrm{E}{-}1$} \\
\bottomrule
\end{tabular}}
\vspace*{-0.5\baselineskip}
\end{table}


Finally, we compare our bias-based neuron-importance score with four standard activation-based criteria.
Recall that the bias-based score for neuron $(\ell,n)$ is $I^{(\ell)}_n$ in~\eqref{eq:bias-importance}.
For the observational dataset in~\eqref{eq:observations}, let
$\mathcal{X}=\{\xi_i\}_{i=1}^N$ be the corresponding set of network inputs.
For a hidden layer $\ell\in\{1,\dots,L\}$ and neuron index $n\in\{1,\dots,N_\ell\}$, let
$a^{(\ell)}_n(\xi)$ denote the post-activation output of neuron $(\ell,n)$ evaluated at input $\xi\in\mathcal{X}$.
We consider the following activation-based importance scores:
\vspace*{-0.4\baselineskip}
\begin{align}
I^{(\ell)}_{n,\mathrm{Freq}}(\mathcal{X})
&:=
\frac{1}{N}\,
\#\bigl\{ \xi\in \mathcal{X} : a_n^{(\ell)}(\xi) > 0 \bigr\}, \qquad
I^{(\ell)}_{n,\mathrm{Abs}}(\mathcal{X})
:=
\frac{1}{N} \sum_{\xi \in \mathcal{X}} \bigl|a_n^{(\ell)}(\xi)\bigr|, \\
I^{(\ell)}_{n,\mathrm{RMS}}(\mathcal{X})
&:=
\sqrt{\frac{1}{N} \sum_{\xi \in \mathcal{X}} \bigl(a_n^{(\ell)}(\xi)\bigr)^2}, \qquad
I^{(\ell)}_{n,\mathrm{Std}}(\mathcal{X})
:=
\sqrt{\frac{1}{N} \sum_{\xi \in \mathcal{X}}
\bigl(a_n^{(\ell)}(\xi) - \overline{a}_n^{(\ell)}\bigr)^2},
\vspace*{-1.0\baselineskip}
\end{align}
where $\#\{\cdot\}$ denotes set cardinality and
\(
\overline{a}_n^{(\ell)} := \frac{1}{N} \sum_{\xi\in\mathcal{X}} a_n^{(\ell)}(\xi)
\)
is the empirical mean activation of neuron $(\ell,n)$ on $\mathcal{X}$.

Table~\ref{tab:criteria_l2re} compares the bias-based importance with these four
activation-based criteria under identical pruning ratios and retention levels.
The bias-based criterion achieves the most favorable trade-off between sparsity
and accuracy across the four problems; in particular, magnitude-based criteria
can lead to substantial degradation on NSInv, whereas the bias-based strategy
remains robust. This comparison suggests that incorporating bias information
into the importance scores provides a more reliable signal for pruning in the
presence of noisy observations.

\begin{table}[htbp]
\centering
\caption{Effect of neuron-importance criteria on L2RE (mean $\pm$ std) across four PDEs.}
\label{tab:criteria_l2re}

\setlength{\tabcolsep}{3pt}
\renewcommand{\arraystretch}{1.0}
\scalebox{0.8}{
\begin{tabular}{c|cccc}
\toprule
\textbf{Criterion} & \textbf{Heat} & \textbf{Wave} & \textbf{PInv} & \textbf{NSInv} \\
\midrule
Bias &
$3.26\mathrm{E}{-}2$ & $\mathbf{8.88\mathrm{E}{-}2}$ & $\mathbf{1.82\mathrm{E}{-}2}$ & $\mathbf{6.44\mathrm{E}{-}1}$ \\
& {\scriptsize$\pm1.56\mathrm{E}{-}4$} & {\scriptsize$\pm4.38\mathrm{E}{-}4$} &
  {\scriptsize$\pm2.17\mathrm{E}{-}3$} & {\scriptsize$\pm2.67\mathrm{E}{-}1$} \\
\midrule
Abs &
$2.59\mathrm{E}{-}2$ & $1.01\mathrm{E}{-}1$ & $3.52\mathrm{E}{-}2$ & $2.01\mathrm{E}{+}0$ \\
& {\scriptsize$\pm1.52\mathrm{E}{-}4$} & {\scriptsize$\pm7.77\mathrm{E}{-}4$} &
  {\scriptsize$\pm3.14\mathrm{E}{-}3$} & {\scriptsize$\pm6.40\mathrm{E}{-}1$} \\
\midrule
RMS &
$\mathbf{2.19\mathrm{E}{-}2}$ & $8.21\mathrm{E}{-}1$ & $3.27\mathrm{E}{-}2$ & $1.48\mathrm{E}{+}0$ \\
& {\scriptsize$\pm3.80\mathrm{E}{-}5$} & {\scriptsize$\pm3.14\mathrm{E}{-}3$} &
  {\scriptsize$\pm9.36\mathrm{E}{-}3$} & {\scriptsize$\pm5.47\mathrm{E}{-}2$} \\
\midrule
Freq &
$6.92\mathrm{E}{-}2$ & $8.96\mathrm{E}{-}2$ & $2.27\mathrm{E}{-}2$ & $1.02\mathrm{E}{+}0$ \\
& {\scriptsize$\pm3.26\mathrm{E}{-}4$} & {\scriptsize$\pm6.36\mathrm{E}{-}4$} &
  {\scriptsize$\pm6.36\mathrm{E}{-}3$} & {\scriptsize$\pm4.79\mathrm{E}{-}1$} \\
\midrule
Std &
$1.01\mathrm{E}{-}1$ & $9.05\mathrm{E}{-}2$ & $3.22\mathrm{E}{-}2$ & $2.81\mathrm{E}{+}0$ \\
& {\scriptsize$\pm8.28\mathrm{E}{-}4$} & {\scriptsize$\pm4.58\mathrm{E}{-}4$} &
  {\scriptsize$\pm4.12\mathrm{E}{-}3$} & {\scriptsize$\pm9.58\mathrm{E}{-}1$} \\
\bottomrule
\end{tabular}
}
\vspace*{-0.5\baselineskip}
\end{table}

\subsection{Sensitivity to noise and pruning hyperparameters}
\label{subsec:noise-sensitivity}
\label{subsec:sensitivity-prune-retain}

We finally study the sensitivity of P-PINN to observational noise and pruning
hyperparameters. Table~\ref{tab:heat_wave_noise} reports L2RE for the heat and
wave data assimilation problems as the noise standard deviation is varied.
Across all noise levels, P-PINN maintains substantially lower error than the
baseline PINN. As the noise standard deviation increases from $0.25$ to $1.0$,
the baseline error grows rapidly, whereas P-PINN degrades more gradually,
which demonstrates robustness to increasingly severe corruption of the data.

\begin{table}[htbp]
\centering
\caption{Mean $\pm$ std of L2RE under varying noise levels for heat and wave data assimilation.}
\scalebox{0.8}{
\setlength{\tabcolsep}{3pt}
\begin{tabular}{c|cc|cc}
\toprule
\textbf{Noise std} & \multicolumn{2}{c|}{\textbf{Heat}} & \multicolumn{2}{c}{\textbf{Wave}} \\
\midrule
                  & PINN & P-PINN & PINN & P-PINN \\
\midrule
\multirow{2}{*}{0.25}
  & $5.25\mathrm{E}{-}1$ & $3.71\mathrm{E}{-}2$ & $4.96\mathrm{E}{-}2$ & $9.90\mathrm{E}{-}3$ \\
  & {\scriptsize$\pm1.99\mathrm{E}{-}2$} & {\scriptsize$\pm9.20\mathrm{E}{-}3$} & {\scriptsize$\pm1.42\mathrm{E}{-}2$} & {\scriptsize$\pm4.20\mathrm{E}{-}3$} \\
\midrule
\multirow{2}{*}{0.50}
  & $1.06\mathrm{E}{+}0$ & $4.89\mathrm{E}{-}2$ & $1.94\mathrm{E}{-}1$ & $2.89\mathrm{E}{-}2$ \\
  & {\scriptsize$\pm1.02\mathrm{E}{+}0$} & {\scriptsize$\pm8.80\mathrm{E}{-}3$} & {\scriptsize$\pm1.29\mathrm{E}{-}1$} & {\scriptsize$\pm1.09\mathrm{E}{-}2$} \\
\midrule
\multirow{2}{*}{0.75}
  & $1.64\mathrm{E}{+}0$ & $1.77\mathrm{E}{-}1$ & $1.30\mathrm{E}{-}1$ & $2.23\mathrm{E}{-}2$ \\
  & {\scriptsize$\pm5.05\mathrm{E}{-}1$} & {\scriptsize$\pm1.91\mathrm{E}{-}1$} & {\scriptsize$\pm4.87\mathrm{E}{-}2$} & {\scriptsize$\pm4.50\mathrm{E}{-}3$} \\
\midrule
\multirow{2}{*}{1.00}
  & $2.70\mathrm{E}{+}0$ & $3.14\mathrm{E}{-}1$ & $3.90\mathrm{E}{-}1$ & $5.41\mathrm{E}{-}2$ \\
  & {\scriptsize$\pm1.39\mathrm{E}{+}0$} & {\scriptsize$\pm2.53\mathrm{E}{-}1$} & {\scriptsize$\pm1.78\mathrm{E}{-}1$} & {\scriptsize$\pm2.96\mathrm{E}{-}2$} \\
\bottomrule
\end{tabular}}
\label{tab:heat_wave_noise}
\end{table}

\begin{figure}[ht]
    \centering
    \includegraphics[width=0.8\linewidth]{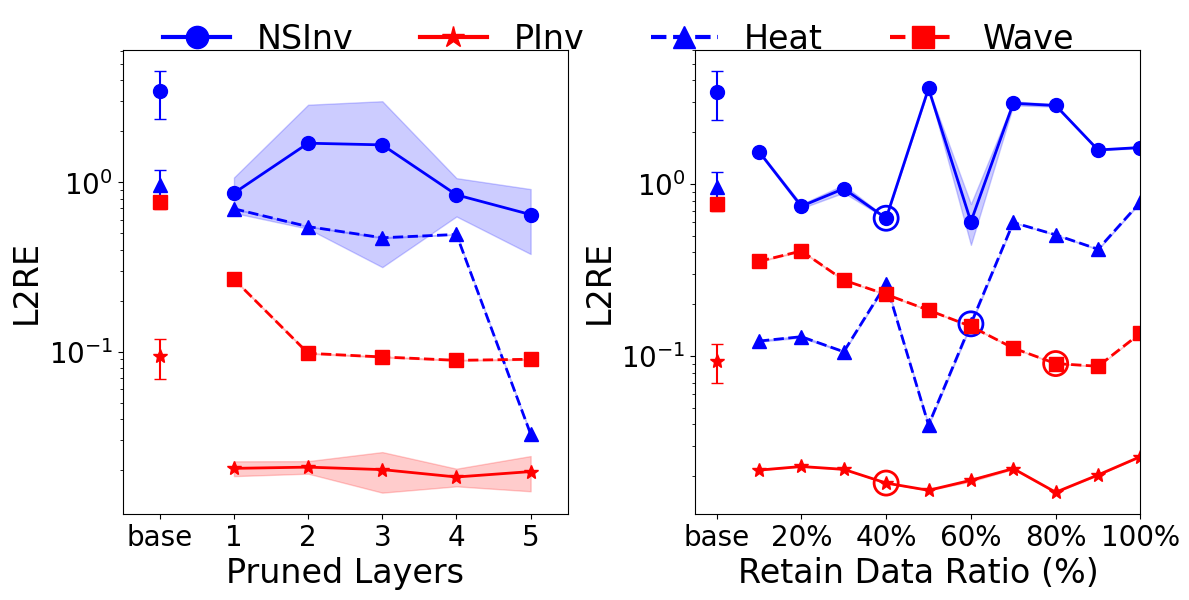}
    \caption{L2RE for four inverse problems (NSInv, PInv, Heat, Wave) as a function of (left) the number of pruned layers and (right) the percentage of retained data (with five layers pruned). The label “base” denotes the baseline PINN trained on the full noisy dataset. Solid lines show means over three runs; shaded bands indicate $\pm 1$ standard deviation. Hollow markers in the right panel highlight the retention ratios used in the main experiments.}
    \label{fig:prune-strategy}
\end{figure}

Figure~\ref{fig:prune-strategy} examines the dependence of P-PINN performance on the number of pruned layers and on the retention ratio $\rho$ for four inverse
problems (NSInv, PInv, Heat, Wave). In the left panel, pruning any positive number of hidden layers already improves over the unpruned baseline PINN (“base”) on all problems. For PInv, the L2RE curves are relatively flat once pruning
is enabled, with three to four pruned layers giving the best average performance. For NSInv, the behavior is mildly non-monotone, but moderate-to-aggressive pruning
(four to five layers) yields the lowest L2RE, and all pruned configurations remain
substantially better than the baseline. The right panel shows that P-PINN is
robust to the choice of data-retention ratio $\rho$: over a broad range of
$\rho$, even fairly aggressive pruning of the observational set continues to
improve upon training on the full noisy dataset, with problem-dependent optimal
choices around intermediate retention levels.

Together, these experiments indicate that P-PINN delivers substantial accuracy
gains while remaining robust both to the level of observational noise and to
moderate misspecification of pruning hyperparameters.


\section{Conclusions}\label{sec:conclusion}

We have proposed P-PINN, a pruning-based framework for robust physics-informed learning from noisy observations in PDE-constrained inverse problems. Starting from a standard PINN trained on the full noisy dataset, P-PINN partitions the observations into retained and forget subsets, prunes hidden neurons according to a bias-based importance score, and performs a short fine-tuning stage on the retained data. The procedure is architecture-agnostic and requires only black-box access to the underlying PINN training routine. Numerical experiments on nine benchmark problems show that P-PINN consistently improves reconstruction accuracy over both the baseline PINN and two natural post hoc baselines (retraining and naive fine-tuning on the retained data). Across a range of noise levels and PDE types, P-PINN reduces relative errors and high-frequency spectral errors while incurring a computational cost that is comparable to naive fine-tuning and substantially lower than full retraining. Our study further indicates that iterative pruning with recomputed importance scores and the proposed bias-based neuron metric yield a favorable accuracy–sparsity trade-off and are robust to variations in pruning depth and data retention ratio. 

Despite its robust performance, P-PINN has several limitations that motivate future research. First, the theoretical foundations of machine unlearning remain underdeveloped: rigorous analyses of unlearning-induced parameter dynamics, stability guarantees, and convergence behavior are still largely absent from the literature. Second, developing adaptive pruning strategies that dynamically recalibrate neuron importance could further enhance noise resilience while reducing computational overhead. Third, integrating Bayesian uncertainty quantification—via stochastic pruning or variational inference—would yield principled estimates of epistemic and aleatoric uncertainties, improving interpretability and enabling rigorous error bounds. Future work will also broaden P-PINN’s scope by (i) extending the selective pruning and fine-tuning paradigm to diverse neural architectures and data-driven pre-training regimes and (ii) evaluating its efficacy on complex multi-physics PDE systems with heterogeneous geometries and multi-scale phenomena. These efforts will strengthen the theoretical foundations of P-PINN and expand its applicability for real-world, noise-affected inverse PDE modeling.

\bibliographystyle{siamplain}
\bibliography{ex_article}

\end{document}